\documentclass[10pt,twocolumn,letterpaper]{article}
\ifdefined\XeTeXversion
  \usepackage[OT1]{fontenc}
\fi
\usepackage[pagenumbers]{cvpr}
\usepackage{graphicx}
\usepackage{caption}
\usepackage{amsmath,amssymb}
\usepackage{fix-cm}
\usepackage{booktabs}
\usepackage{xspace}
\usepackage{multirow}
\usepackage{tabularx}
\newcommand{\dE}{\ensuremath{\Delta E_{00}}\xspace}

\makeatletter
\newcommand{\resultgroupcolor}[3]{%
  \multicolumn{#1}{@{}l@{}}{%
    \rlap{\smash{\color[HTML]{#2}%
      \rule[-\dp\@arstrutbox]{\textwidth}{\dimexpr\ht\@arstrutbox+\dp\@arstrutbox\relax}}}%
    \hspace{2pt}\emph{#3}} \\}
\makeatother

\definecolor{cvprblue}{rgb}{0.21,0.49,0.74}
\usepackage[pagebackref,breaklinks,colorlinks,allcolors=cvprblue]{hyperref}
\def\confName{CVPR}
\def\confYear{2027}
\title{\texorpdfstring{\textcolor[HTML]{A85F65}{L}\textcolor[HTML]{B17B45}{U}\textcolor[HTML]{9A8637}{T}\textcolor[HTML]{6F8B58}{B}\textcolor[HTML]{508D8C}{e}\textcolor[HTML]{5E7DA6}{n}\textcolor[HTML]{8173A5}{c}\textcolor[HTML]{986EA0}{h}}{LUTBench}: Benchmarking Text-Driven Photographic Look Transfer}
\author{%
Chuanzhi Xu\textsuperscript{1,*,$\dagger$}\quad
Boyu Wei\textsuperscript{1,*}\quad
Chengkun Yue\textsuperscript{2}\quad
Xuanhua Yin\textsuperscript{1}\\
Yiyang Li\textsuperscript{1}\quad
Chupeng Liu\textsuperscript{1}\quad
Zihan Deng\textsuperscript{3}\quad
Weidong Cai\textsuperscript{1}\\[3pt]
{\normalsize\textsuperscript{1}The University of Sydney\quad
\textsuperscript{2}Indiana University\quad
\textsuperscript{3}The University of Hong Kong\quad\href{https://github.com/LouckXu/LUTBench}{\raisebox{-0.12em}{\includegraphics[height=0.9em]{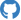}}\hspace{0.3em}Project Page}}}
\hypersetup{pdftitle={LUTBench: Benchmarking Text-Driven Photographic Look Transfer},pdfauthor={Chuanzhi Xu, Boyu Wei, Chengkun Yue, Xuanhua Yin, Yiyang Li, Chupeng Liu, Zihan Deng, Weidong Cai}}
\makeatletter
\renewcommand{\@maketitle}{\begin{center}{\Large\bfseries\@title\par}\vspace{14pt}{\large\@author\par}\end{center}\vspace{10pt}}
\makeatother

\makeatletter
\let\paperfulltitle\@title
\makeatother

\newcommand{\estimateci}[2]{\begin{tabular}[t]{@{}c@{}}#1\\[-2pt]{\scriptsize #2}\end{tabular}}

\makeatletter
\g@addto@macro\@maketitle{%
\begin{minipage}{\textwidth}
\centering
\includegraphics[width=\textwidth]{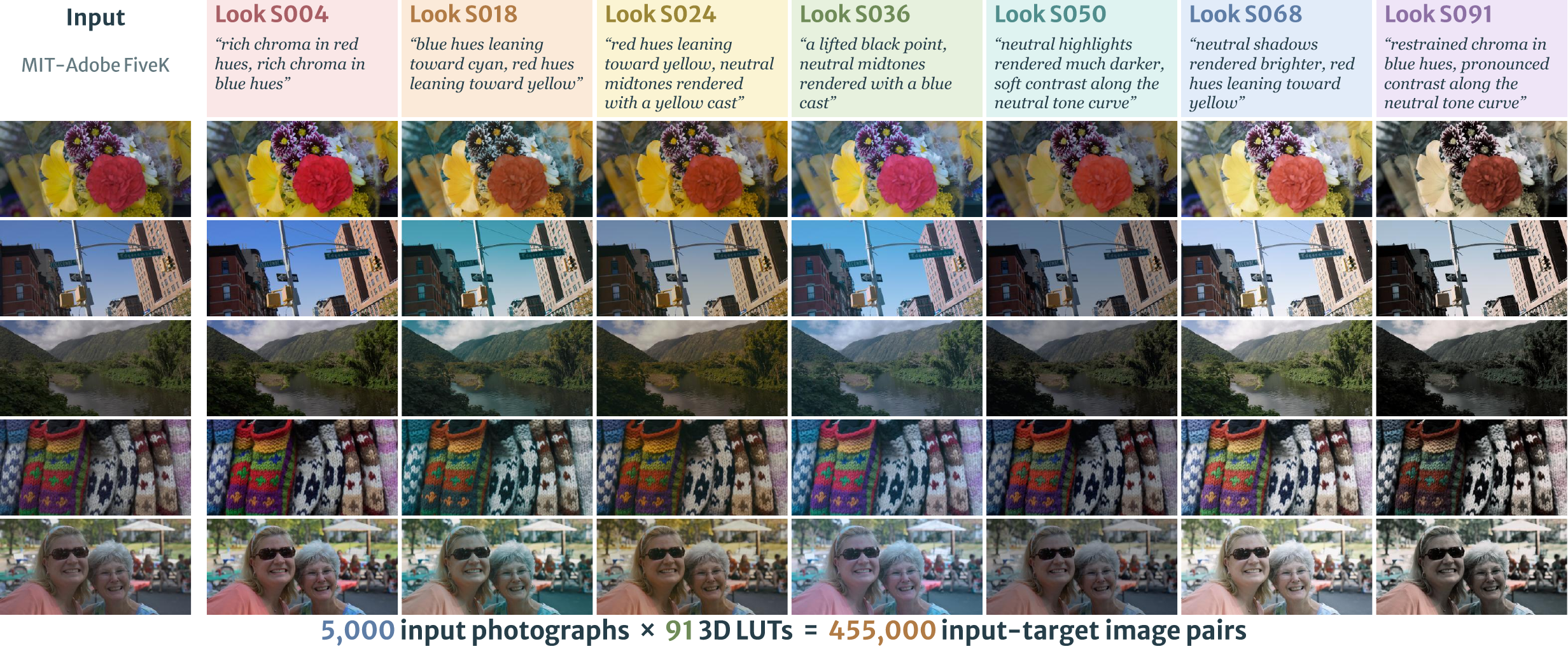}
\captionsetup{hypcap=false}
\captionof{figure}{Examples from LUTBench. Each of the five rows shows an input photograph and its canonical target images for seven looks. The text above each target column contains two verbatim clauses from the corresponding measurement-based look description.}
\label{fig:overview}
\end{minipage}\par
\vskip 8pt plus 2pt minus 2pt
}
\makeatother

\begin{document}
\raggedbottom
\crefname{appendix}{Appendix}{Appendices}
\Crefname{appendix}{Appendix}{Appendices}
\maketitle
\begingroup
\renewcommand{\thefootnote}{\fnsymbol{footnote}}
\footnotetext[1]{Equal contribution.}
\footnotetext[2]{Corresponding author: \href{mailto:chuanzhi.xu@sydney.edu.au}{\texttt{chuanzhi.xu@sydney.edu.au}}.}
\endgroup

\AddToHookNext{env/abstract/begin}{\setlength{\parskip}{0pt}}
\begin{abstract}
Photographic color grading shapes the tone and color of a photograph to achieve a desired look. Text-driven photographic look transfer lets users specify these adjustments in natural language, with the goal of reproducing the requested look across photographs while preserving their content. However, prior research on text-driven color grading uses different datasets, text instructions, target definitions, and metrics. These differences make it difficult to compare how faithfully the methods reproduce the same look. To compare methods under a shared protocol, we introduce LUTBench. We collect 91 three-dimensional lookup tables (3D LUTs) and define one photographic look per LUT with a specified input conversion. Applying these looks to 5,000 MIT-Adobe FiveK photographs produces 455,000 input-target image pairs with reproducible target images. For each look, we use templates to turn measured tone and color changes on fixed color samples into candidate clauses. We use training image measurements to verify applicable clauses before composing the text description. The evaluation measures target fidelity, tests sensitivity to predefined input conversions, and separates generalization to unseen looks from reproduction of seen looks on new photographs. Across 16 methods, stronger text alignment does not guarantee target fidelity. However, most supervised variants improve fidelity on both seen and unseen looks.
\end{abstract}

\section{Introduction}
\label{sec:intro}

To apply the same look to a photo series, a photographer might write, \textit{``Warm the highlights, reduce chroma in blue regions, and lift the shadows.''} A text-driven color grading method must reproduce these adjustments as the subjects and initial colors vary across photographs. Text-conditioned tone adjustment~\cite{lee2024cliptone,mayer2025lcit}, instruction-based image editing~\cite{brooks2023instructpix2pix,li2025superedit}, and retouching agents~\cite{dutt2025monetgpt,veraretouch2026} can all turn such requests into edited photographs.

However, prior research uses different test protocols, with variation in datasets, text instructions, target definitions, and metrics. For text-conditioned tone adjustment, one protocol measures text alignment and structure preservation using manually collected tone descriptions and separately compares edited images with filter-rendered targets~\cite{lee2024cliptone}. In general instruction editing, a dataset provides manually annotated targets for a broader range of edits~\cite{zhang2023magicbrush}. Work on instruction-guided color grading measures perceptual similarity to LUT-rendered targets and reports model-based preference rankings~\cite{ma2026acetone}. Some prior research reports results on study-specific test sets without making the full evaluation data publicly available~\cite{conde2025pixtalk,veraretouch2026,ma2026acetone,wu2026instantretouch,wu2026retouchiq}. Because these protocols measure different aspects of edit quality, direct comparison requires a shared benchmark with the same look, input photographs, target definition, and metrics.

When text-driven color grading methods receive the same text description for different photographs, assessing how faithfully they reproduce the requested look requires a target image for each input. Even when two outputs match the requested direction of an adjustment, the magnitude of the change can vary widely. If a fixed color transformation defines the requested look, applying it to each input photograph gives the corresponding target image.

In this paper, we introduce LUTBench, a benchmark for text-driven photographic look transfer. We define each look with a three-dimensional lookup table (3D LUT) and a canonical input conversion that maps each photograph to the LUT's selected input encoding. Applying the LUT to the converted photograph renders its canonical target image, as shown in \cref{fig:overview}. In each comparison, we use a shared evaluation protocol with the same photographs and target image sets for all methods and the same description per look for text-driven methods.

Our contributions can be summarized as follows:
\begin{itemize}
\item We collect 91 3D LUTs and define one photographic look per LUT using a canonical input conversion. Applying these looks to 5,000 photographs produces 455,000 input-target image pairs. We construct measurement-based look descriptions from quantified tone and color changes.
\item We define a protocol that keeps similar LUT transformations in the same partition and uses reproducible target images to assess seen-look reproduction and unseen-look generalization. Paired comparisons of canonical error and minimum error over candidate input conversions test sensitivity to the conversion settings.
\item We evaluate 16 methods from prior research through zero-shot text-driven configurations, supervised variants, and a reference-guided configuration. We compare target fidelity and assess its link to semantic alignment.
\end{itemize}

\section{Related Work}
\label{sec:related}
\noindent\textbf{Text-Guided Image Editing.}
CLIPtone~\cite{lee2024cliptone} adapts a LUT-based photo enhancer using CLIP~\cite{radford2021clip} text directions, and \mbox{CLIPtone-GO}~\cite{merothiya2026cliptonego} refines the optimization to reduce artifacts. LCIT~\cite{mayer2025lcit} fits a color transformation matrix. In contrast, InstantRetouch~\cite{wu2026instantretouch} predicts photometric transformations in bilateral space. CLIPstyler~\cite{kwon2022clipstyler}, SpectralCLIP~\cite{xu2024spectralclip}, FastCLIPstyler~\cite{suresh2024fastclipstyler}, and ZeCon~\cite{yang2023zecon} address text-guided style transfer. InstructPix2Pix~\cite{brooks2023instructpix2pix}, SuperEdit~\cite{li2025superedit}, StyleBooth~\cite{han2024stylebooth}, SPL~\cite{gong2026spl}, and Qwen-Image-Edit~\cite{wu2025qwenimage} support broader appearance edits. T2ONet~\cite{shi2021t2onet} plans sequences of global adjustments. Retouching agents such as IEA~\cite{iea2026}, MonetGPT~\cite{dutt2025monetgpt}, and VeraRetouch~\cite{veraretouch2026} infer editing parameters. LUTBench compares representative methods across these categories under a shared protocol for text-driven photographic look transfer.

\noindent\textbf{LUT-Based and Reference-Guided Color Grading.}
Image-adaptive LUTs~\cite{zeng2020lut} and adaptive sampling intervals~\cite{yang2022adaint} support efficient photo enhancement. NILUT~\cite{conde2024nilut} represents transformations implicitly, and PG-IA-NILUT~\cite{pgianilut2024} exposes interpretable filter controls. Reference-guided methods match color statistics~\cite{reinhard2001color} or predict spatially adaptive LUTs~\cite{gong2025salut}. AceTone~\cite{ma2026acetone} and StatLUT~\cite{wang2026statlut} extend LUT prediction to text conditioning. LUTBench fixes both the LUT and its canonical input conversion, so each canonical target image is independent of the evaluated method.

\noindent\textbf{Color Grading and Retouching Benchmarks.}
FiveK~\cite{bychkovsky2011fivek} and PPR10K~\cite{liang2021ppr10k} provide expert-retouched targets. MA5k-Req~\cite{shi2021t2onet} and iRetouch~\cite{wu2026instantretouch} pair individual photo edits with language requests. In contrast, PST50~\cite{gong2025salut} and the Retouch Transfer Dataset~\cite{elezabi2026inretouch} use image references to specify the look. AceTone-Bench~\cite{ma2026acetone} provides LUT-rendered targets for both instruction- and reference-guided grading. However, its instruction subset uses descriptions generated for individual image-LUT pairs, which limits what aggregate scores reveal about fidelity to the same described look across photographs. LUTBench pairs each look with one description derived from measured tone and color changes and reuses that description across a common set of photographs with corresponding targets. Keeping similar looks in the same partition lets LUTBench assess seen-look reproduction and unseen-look generalization separately.

\section{Constructing LUTBench}
\label{sec:benchmark}
\begin{figure*}[t]
\centering
\includegraphics[width=\textwidth]{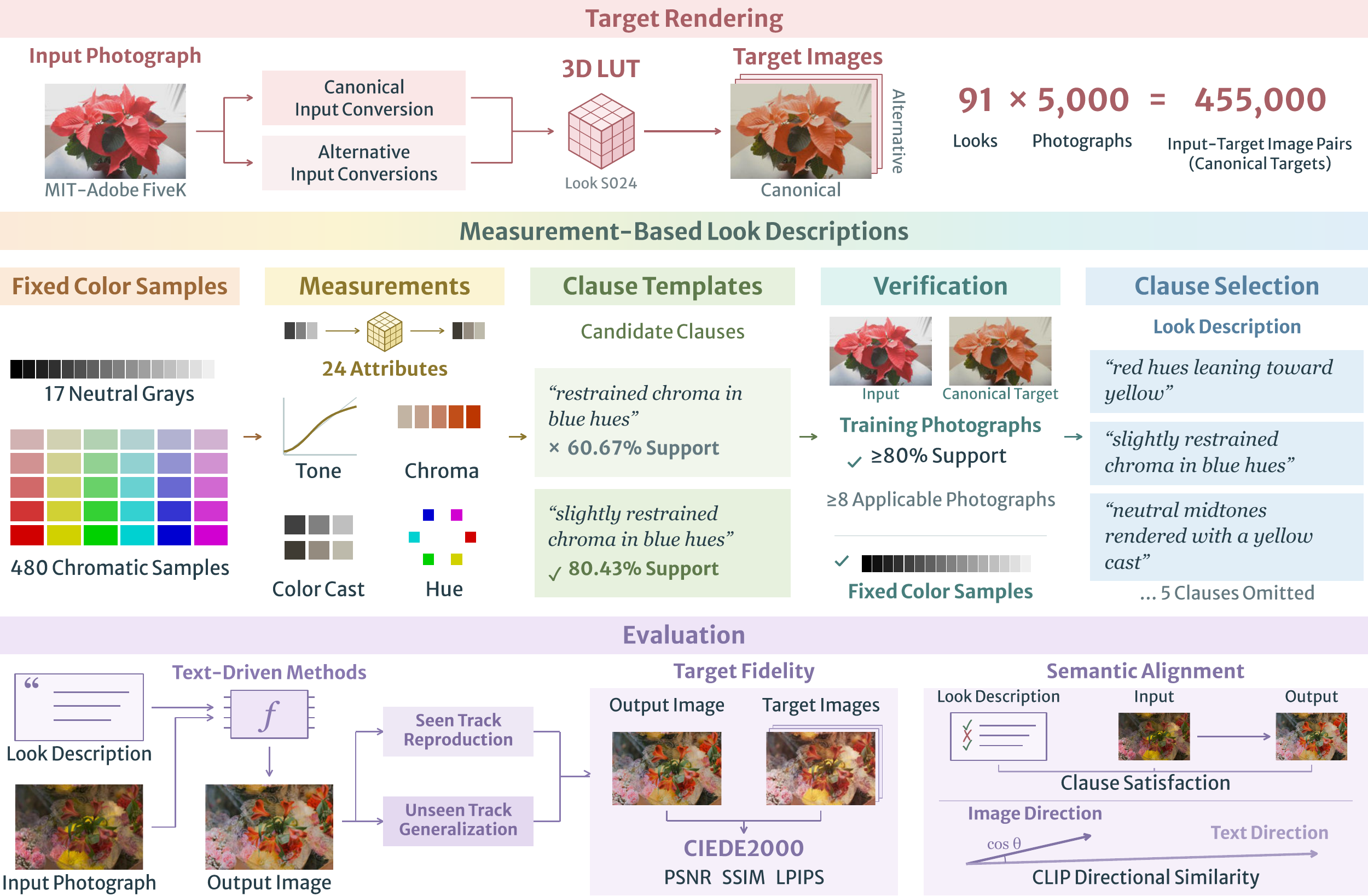}
\caption{Overview of LUTBench. LUTs render paired target images. Measurements guide clause generation and verification to construct look descriptions. The paired images and descriptions support evaluation of target fidelity and semantic alignment.}
\label{fig:construction}
\end{figure*}
\subsection{Task and Reproducible Target Images}
\Cref{fig:construction} provides an overview of LUTBench. Let $\hat y_{il}$ denote the output image for input photograph $x_i$ under look $l$. For a text-driven configuration $f$ with look description $d_l$, the output is $\hat y_{il}=f(x_i,d_l)$. A look consists of a three-dimensional lookup table (3D LUT) $T_l$ and a canonical input conversion $r_l$. The canonical target image for input photograph $x_i$ and look $l$ is:
\begin{equation}
 y_{il}=Q_{16}\!\left(T_l(r_l(x_i))\right),
 \label{eq:target}
\end{equation}
where $Q_{16}$ denotes 16-bit quantization of each RGB channel. The canonical input conversion $r_l$ maps the input photograph to the LUT's input encoding, and the rendered output is interpreted as display sRGB. The canonical target image shows the effect of the LUT's RGB mapping on the input photograph. The look description summarizes the measured changes in tone and color.

LUTBench uses all 5,000 RAW photographs from MIT-Adobe FiveK~\cite{bychkovsky2011fivek}, developed with the same neutral RAW-processing settings. Applying the 91 looks (S001 to S091) produces 455,000 input-target image pairs. Photograph preparation and partition details are in \cref*{sec:supp_photographs}. LUT grids and input conversions are specified in \cref*{sec:supp_conversions}.

\begin{figure}[t]
\centering
\includegraphics[width=\columnwidth]{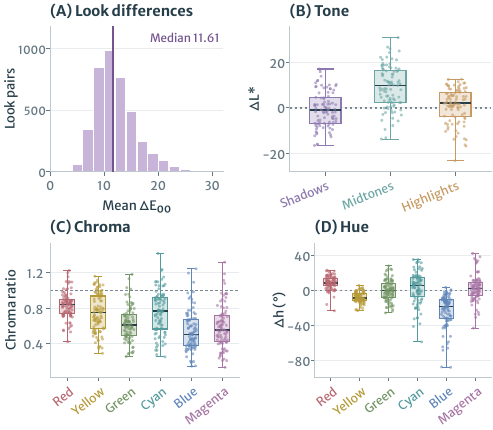}
\caption{Color and tone variation across 91 looks. Panel A marks the median pairwise mean CIEDE2000. Dots denote looks, boxes show medians and interquartile ranges, and whiskers span the full range. Dashed lines mark unchanged attribute values.}
\label{fig:look_diversity}
\end{figure}

Regarding color and tone variation, \cref{fig:look_diversity} shows that the 91 looks include both lightening and darkening within each tonal region, with chroma and hue changes that vary across hue sectors. Reproducing these looks requires methods to match both the direction and magnitude of the changes. Measurement statistics are in \cref*{sec:supp_descriptions}.

\begin{figure}[t]
\centering
\includegraphics[width=\columnwidth]{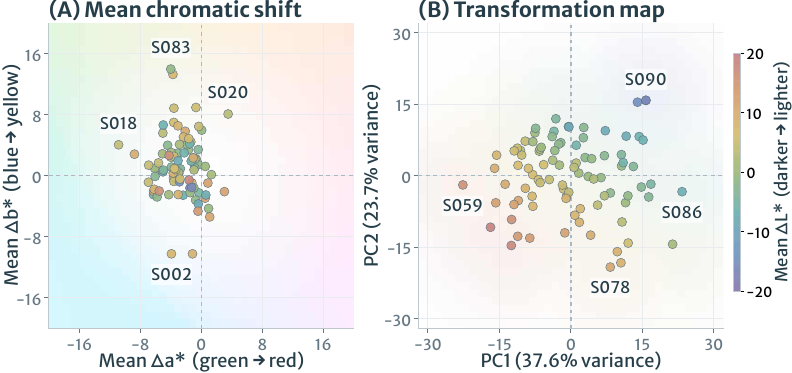}
\caption{Each point represents one of 91 looks, measured on the same 497 fixed color samples. Point colors encode mean $\Delta L^*$. Backgrounds indicate chromatic directions in A and smoothed local lightness changes in B.}
\label{fig:look_color_space}
\end{figure}

\Cref{fig:look_color_space} provides two complementary views of the look collection. Mean chromatic shifts span blueward and yellowward directions. However, averaging can hide adjustments that differ across input colors. We therefore also project the complete CIELAB change vectors using principal component analysis (PCA). Its first two components explain 61.3\% of between-look variance, summarizing transformation differences beyond mean color shifts. Construction details are in \cref*{sec:supp_descriptions}.

\subsection{Measurement-Based Look Descriptions}
LUTBench look descriptions express four types of adjustment: tone, chroma, color cast (color introduced into initially neutral grays), and hue. We measure each LUT under its canonical input-conversion settings using the same predefined sRGB samples: 17 neutral grays and 480 chromatic colors. These measurements define 24 numerical attributes of the LUT's tone and color effects. They quantify black and white points (the output lightness of black and white samples), contrast, color casts in neutral shadows, midtones, and highlights, and chroma and hue changes within six hue sectors. Tone attributes for shadows, midtones, and highlights measure output-minus-input lightness. Predefined templates use these measurements and numerical thresholds to generate candidate clauses, each a short statement specifying an adjustment's region, direction, and degree. Examples include \textit{``neutral shadows rendered with a subtle blue cast''} and \textit{``restrained chroma in blue hues.''}

However, photographs differ in their color and tonal content. We therefore check the clauses whose adjustments can be measured by comparing training photographs with their canonical target images. For candidate clause $c$, let $a_{ilc}$ be one when its attribute is measurable for photograph $i$ under look $l$, and zero otherwise. Let $v_{ilc}$ indicate whether the measured adjustment satisfies the clause. The clause support rate is the fraction of measurable training photographs whose canonical targets satisfy the clause, computed on the training set $\mathcal I_{\mathrm{train}}$:
\begin{equation}
 s_{lc}=\frac{\sum_{i\in\mathcal I_{\mathrm{train}}}a_{ilc}v_{ilc}}
 {\sum_{i\in\mathcal I_{\mathrm{train}}}a_{ilc}}.
 \label{eq:support}
\end{equation}
A candidate is accepted when $s_{lc}\geq0.8$ and its attribute is measurable on at least eight training photographs. Black-point, white-point, and contrast clauses are checked only on fixed color samples, which cover the required tonal range.

We rank verified clauses and combine them into one description per look, limiting the number of clauses from each adjustment type. Ranking uses attribute measurements and clause support rates, with measurements normalized across all 91 looks. We choose the clause selection rule minimizing mean CIEDE2000 across three text-conditioned development models on held-out looks and validation photographs. On validation photographs, clauses checked using image measurements reach 98.8\% agreement in adjustment direction and 93.1\% support for their full numerical conditions, averaged equally across all looks. Templates, thresholds, ranking, and clause-count limits are detailed in \cref*{sec:supp_descriptions}. In the human evaluation in \cref*{sec:supp_human}, 13 raters scored how accurately clauses describe visible input-to-target changes, averaging 3.93 on a five-point scale, with 68.9\% of ratings at least ``mostly matches''.

\subsection{Input-Conversion Settings}
Before a LUT is applied, the input conversion specifies the encoding curve, RGB primaries, and exposure offset. These settings determine the RGB values supplied to the LUT, so changing them can alter the target image rendered from the same photograph. LUTBench fixes a canonical input conversion for each look using format metadata or input-conversion calibration against matched sRGB exports.

For sensitivity analysis, LUTBench renders alternative target images using predefined alternative input conversions while keeping the photograph and LUT fixed. The target-fidelity measures are defined in \cref{sec:target_fidelity}. Conversion settings and sensitivity measurements are in \cref*{sec:supp_conversions}.

\subsection{Look Groups and Partitions}
\label{sec:look_partitions}
\begin{figure}[t]
\centering
\includegraphics[width=\columnwidth]{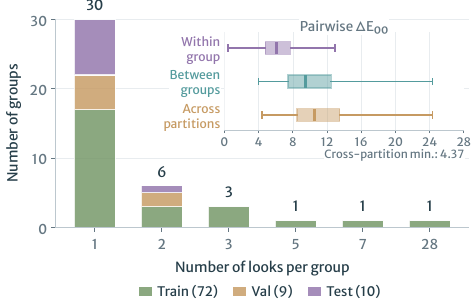}
\caption{The main bar chart shows look-group counts by size and partition. The inset boxplot compares pairwise target color differences within groups, between groups, and across partitions.}
\label{fig:look_groups}
\end{figure}

To reduce similarity between training and evaluation looks, we group LUTs by their color transformations. We connect two looks when the mean CIEDE2000 difference between their canonical target images on a fixed set of training photographs is below 4 \dE. Each connected component defines a look group. The fixed partitions keep all 42 groups intact and contain 72 training looks, 9 validation looks (Val), and 10 test looks (Test). The closest pair across partitions is 4.37 \dE{} apart. \Cref{fig:look_groups} shows group sizes and distances.

\section{Evaluation Protocol}
\label{sec:protocol}
\subsection{Evaluation Tracks}
\label{sec:eval_tracks}
We evaluate seen-look reproduction (Seen track) and unseen-look generalization (Unseen track). The Seen track uses the training descriptions to assess reproduction of the 72 training looks on new photographs. The Unseen track comprises Val, Test and cross-validation (CV). Val and Test use the 9 validation and 10 test looks excluded from training. Each supervised variant uses the same model trained on 72 looks for the Seen track, Val and Test.

CV extends the Unseen track to 40 non-test looks, including the 9 Val looks. CV pools results from five fixed held-out subsets of eight looks. Each evaluation run trains supervised variants on the other 73 non-test looks and evaluates the held-out subset. CV partitions are detailed in \cref*{sec:supp_splits}. CV and Val use all 500 validation photographs. The Seen track and Test use all 500 test photographs.

\begin{table*}[t]
\centering
\caption{Target fidelity and text alignment. \textcolor[HTML]{2C6092}{Blue} headers denote the Seen track, and \textcolor[HTML]{A04444}{red} headers denote the Unseen track (CV, Val and Test). Regional errors, weighted components and text-alignment metrics use CV looks. Header counts indicate the number of looks. $^*$ marks configurations supervised on LUTBench. In the canonical-error columns, $^{-}$ and $^{+}$ indicate supported improvements and deficits relative to the unedited input under the rule in \cref{sec:paired_comparisons}, respectively. Unmarked method comparisons are inconclusive. Weighted components report mean absolute weighted CIEDE2000 terms, with signed lightness and chroma biases in parentheses. Min. error, Clauses and CLIP dir. denote minimum error over candidate input conversions, clause satisfaction and CLIP directional similarity, respectively.}
\label{tab:main}
\fontsize{6.5}{7.5}\selectfont
\setlength{\tabcolsep}{0.85pt}
\setlength{\arrayrulewidth}{0.25pt}
\begin{tabular*}{\textwidth}{@{\extracolsep{\fill}}l|*{4}{c}|*{4}{c}|*{3}{c}|*{3}{c}|*{2}{c}@{}}
\toprule
 & \multicolumn{4}{c|}{Canonical $\Delta E_{00}\downarrow$} & \multicolumn{4}{c|}{\textcolor[HTML]{A04444}{CV fidelity}} & \multicolumn{3}{c|}{\textcolor[HTML]{A04444}{Regional $\Delta E_{00}\downarrow$}} & \multicolumn{3}{c|}{\textcolor[HTML]{A04444}{Weighted components}} & \multicolumn{2}{c}{\textcolor[HTML]{A04444}{Text alignment $\uparrow$}} \\
\cmidrule(lr){2-5}\cmidrule(lr){6-9}\cmidrule(lr){10-12}\cmidrule(lr){13-15}\cmidrule(l){16-17}
Method & \textcolor[HTML]{A04444}{CV (40)} & \textcolor[HTML]{A04444}{Val (9)} & \textcolor[HTML]{A04444}{Test (10)} & \textcolor[HTML]{2C6092}{Seen (72)}
 & \textcolor[HTML]{A04444}{Min. error $\downarrow$} & \textcolor[HTML]{A04444}{PSNR $\uparrow$} & \textcolor[HTML]{A04444}{SSIM $\uparrow$} & \textcolor[HTML]{A04444}{LPIPS $\downarrow$}
 & \textcolor[HTML]{A04444}{Neutral} & \textcolor[HTML]{A04444}{Chrom.} & \textcolor[HTML]{A04444}{Hue-bal.}
 & \textcolor[HTML]{A04444}{$L$ (bias)} & \textcolor[HTML]{A04444}{$C$ (bias)} & \textcolor[HTML]{A04444}{$H$} & \textcolor[HTML]{A04444}{Clauses} & \textcolor[HTML]{A04444}{CLIP dir.} \\
\midrule
Canonical target image & 0.00 & 0.00 & 0.00 & 0.00 & 0.00 & $\infty$ & 1.000 & 0.000 & 0.00 & 0.00 & 0.00 & 0.0 (0.0) & 0.0 (0.0) & 0.0 & 94\% & 0.036 \\
Unedited input & 8.65 & 9.38 & 10.26 & 7.66 & 7.78 & 22.62 & 0.873 & 0.109 & 8.11 & 9.84 & 10.04 & 5.7 (-4.5) & 3.1 (-1.7) & 3.7 & 7\% & - \\ 
\resultgroupcolor{17}{FAE7E7}{CLIP-guided tone}
CLIPtone~\cite{lee2024cliptone} & 13.18$^{+}$ & 13.80$^{+}$ & 12.77 & 14.08$^{+}$ & 12.57 & 21.64 & 0.773 & 0.185 & 12.87 & 13.34 & 13.39 & 4.8 (-0.6) & 6.9 (+5.5) & 8.2 & 41\% & 0.099 \\ 
LCIT~\cite{mayer2025lcit} & 10.10$^{+}$ & 10.77$^{+}$ & 11.15$^{+}$ & 9.08$^{+}$ & 9.30 & 22.18 & 0.861 & 0.144 & 8.62 & 13.47 & 12.80 & 5.8 (-4.6) & 3.4 (-1.1) & 5.6 & 19\% & 0.083 \\ 
\resultgroupcolor{17}{FBEAD9}{Text style transfer}
CLIPstyler~\cite{kwon2022clipstyler} & 22.64$^{+}$ & 23.07$^{+}$ & 23.68$^{+}$ & 22.22$^{+}$ & 22.05 & 15.39 & 0.426 & 0.451 & 21.15 & 26.34 & 27.23 & 10.3 (+4.4) & 10.8 (+9.9) & 14.2 & 45\% & 0.130 \\ 
FastCLIPstyler~\cite{suresh2024fastclipstyler} & 31.65$^{+}$ & 34.11$^{+}$ & 31.32$^{+}$ & 32.18$^{+}$ & 29.70 & 11.73 & 0.487 & 0.565 & 31.78 & 31.66 & 31.16 & 23.6 (+22.2) & 11.0 (+9.9) & 12.9 & 55\% & 0.092 \\ 
SpectralCLIP~\cite{xu2024spectralclip} & 30.20$^{+}$ & 31.21$^{+}$ & 30.57$^{+}$ & 30.98$^{+}$ & 28.80 & 11.03 & 0.234 & 0.578 & 29.55 & 31.80 & 30.96 & 20.4 (+17.5) & 11.9 (+11.1) & 14.8 & 44\% & 0.105 \\ 
ZeCon~\cite{yang2023zecon} & 14.93$^{+}$ & 15.58$^{+}$ & 15.72$^{+}$ & 14.32$^{+}$ & 14.16 & 16.62 & 0.368 & 0.520 & 13.67 & 17.36 & 16.79 & 9.0 (-6.5) & 6.6 (+2.7) & 6.8 & 37\% & 0.102 \\ 
\resultgroupcolor{17}{FBF4D3}{Instruction editors}
SPL~\cite{gong2026spl} & 18.74$^{+}$ & 19.68$^{+}$ & 18.19$^{+}$ & 18.30$^{+}$ & 17.84 & 16.04 & 0.677 & 0.294 & 17.18 & 21.97 & 21.24 & 12.0 (+0.8) & 6.7 (+3.9) & 9.4 & 43\% & 0.105 \\ 
InstructPix2Pix~\cite{brooks2023instructpix2pix} & 22.20$^{+}$ & 23.89$^{+}$ & 22.76$^{+}$ & 22.37$^{+}$ & 21.57 & 16.19 & 0.579 & 0.362 & 20.69 & 25.38 & 24.53 & 9.8 (+3.0) & 12.2 (+11.6) & 12.6 & 46\% & 0.113 \\ 
SuperEdit~\cite{li2025superedit} & 26.93$^{+}$ & 27.18$^{+}$ & 26.08$^{+}$ & 27.72$^{+}$ & 26.30 & 12.36 & 0.401 & 0.517 & 25.93 & 28.82 & 27.97 & 16.6 (+9.6) & 12.3 (+10.7) & 13.1 & 41\% & 0.095 \\ 
StyleBooth~\cite{han2024stylebooth} & 21.13$^{+}$ & 20.80$^{+}$ & 19.17$^{+}$ & 21.79$^{+}$ & 20.54 & 14.84 & 0.539 & 0.416 & 19.67 & 24.21 & 22.99 & 9.4 (+3.8) & 12.1 (+11.2) & 11.6 & 48\% & 0.148 \\ 
Qwen-Image-Edit-2511~\cite{wu2025qwenimage} & 23.24$^{+}$ & 25.46$^{+}$ & 21.07$^{+}$ & 20.36$^{+}$ & 21.28 & 13.48 & 0.458 & 0.313 & 22.54 & 24.77 & 24.14 & 17.3 (+16.1) & 7.6 (+6.2) & 9.2 & 64\% & 0.097 \\ 
\resultgroupcolor{17}{E7F1DE}{Retouching agents}
IEA~\cite{iea2026} & 14.79$^{+}$ & 15.57$^{+}$ & 16.94$^{+}$ & 17.65$^{+}$ & 14.24 & 18.07 & 0.746 & 0.178 & 14.73 & 14.53 & 15.09 & 11.0 (+6.1) & 4.4 (+2.1) & 5.8 & 37\% & 0.016 \\ 
MonetGPT~\cite{dutt2025monetgpt} & 9.60$^{+}$ & 10.29 & 12.24$^{+}$ & 11.16$^{+}$ & 8.98 & 22.52 & 0.846 & 0.146 & 9.54 & 9.60 & 9.88 & 5.1 (+2.3) & 3.9 (+1.3) & 5.3 & 43\% & 0.036 \\ 
VeraRetouch~\cite{veraretouch2026} & 8.91 & 9.95 & 10.61 & 8.29$^{+}$ & 8.35 & 23.02 & 0.871 & 0.116 & 8.02 & 10.92 & 10.99 & 5.4 (-1.9) & 3.4 (-1.3) & 4.3 & 45\% & 0.023 \\ 
\resultgroupcolor{17}{DFF3F1}{Reference-guided}
AceTone~\cite{ma2026acetone} & 7.82$^{-}$ & 8.76 & 8.59$^{-}$ & 6.44$^{-}$ & 6.82 & 23.67 & 0.905 & 0.090 & 7.22 & 9.26 & 9.44 & 5.2 (-4.5) & 3.0 (-2.1) & 3.2 & 47\% & 0.016 \\ 
\resultgroupcolor{17}{EFE5F7}{Supervised on LUTBench}
CLIPtone$^*$~\cite{lee2024cliptone} & 6.38$^{-}$ & 6.92$^{-}$ & 8.05$^{-}$ & 1.69$^{-}$ & 6.04 & 26.85 & 0.925 & 0.075 & 6.16 & 6.73 & 6.86 & 3.0 (-0.9) & 2.7 (-0.7) & 3.7 & 59\% & 0.028 \\ 
InstructPix2Pix$^*$~\cite{brooks2023instructpix2pix} & 6.51$^{-}$ & 7.16$^{-}$ & 8.03$^{-}$ & 3.77$^{-}$ & 6.13 & 25.56 & 0.809 & 0.106 & 6.17 & 7.06 & 7.00 & 3.3 (-1.4) & 2.8 (-0.9) & 3.6 & 63\% & 0.002 \\ 
SuperEdit$^*$~\cite{li2025superedit} & 6.60$^{-}$ & 7.31$^{-}$ & 8.20$^{-}$ & 3.82$^{-}$ & 6.22 & 25.64 & 0.804 & 0.109 & 6.25 & 7.14 & 7.13 & 3.3 (-1.4) & 2.8 (-0.9) & 3.7 & 61\% & 0.001 \\ 
StyleBooth$^*$~\cite{han2024stylebooth} & 6.10$^{-}$ & 6.68$^{-}$ & 7.69$^{-}$ & 3.05$^{-}$ & 5.65 & 26.75 & 0.884 & 0.084 & 5.76 & 6.72 & 6.70 & 3.0 (-1.4) & 2.6 (-0.8) & 3.4 & 65\% & 0.027 \\ 
VeraRetouch$^*$~\cite{veraretouch2026} & 7.24$^{-}$ & 8.41$^{-}$ & 8.39$^{-}$ & 5.90$^{-}$ & 6.57 & 24.56 & 0.891 & 0.084 & 6.75 & 8.18 & 8.36 & 4.6 (-2.9) & 2.9 (-1.5) & 3.2 & 45\% & 0.019 \\ 
FastCLIPstyler$^*$~\cite{suresh2024fastclipstyler} & 9.61$^{+}$ & 10.48$^{+}$ & 11.47$^{+}$ & 8.63$^{+}$ & 9.15 & 21.37 & 0.759 & 0.247 & 8.61 & 11.77 & 11.25 & 5.9 (-1.9) & 3.7 (-1.6) & 4.6 & 55\% & 0.028 \\ 
PG-IA-NILUT$^*$~\cite{pgianilut2024} & 6.18$^{-}$ & 7.51$^{-}$ & 8.33$^{-}$ & 4.92$^{-}$ & 5.78 & 27.07 & 0.888 & 0.084 & 5.88 & 6.91 & 7.02 & 2.8 (-2.0) & 2.7 (-1.2) & 3.6 & 49\% & 0.019 \\ 
\bottomrule
\end{tabular*}
\end{table*}

\begin{figure*}[t]
\centering
\includegraphics[width=\textwidth]{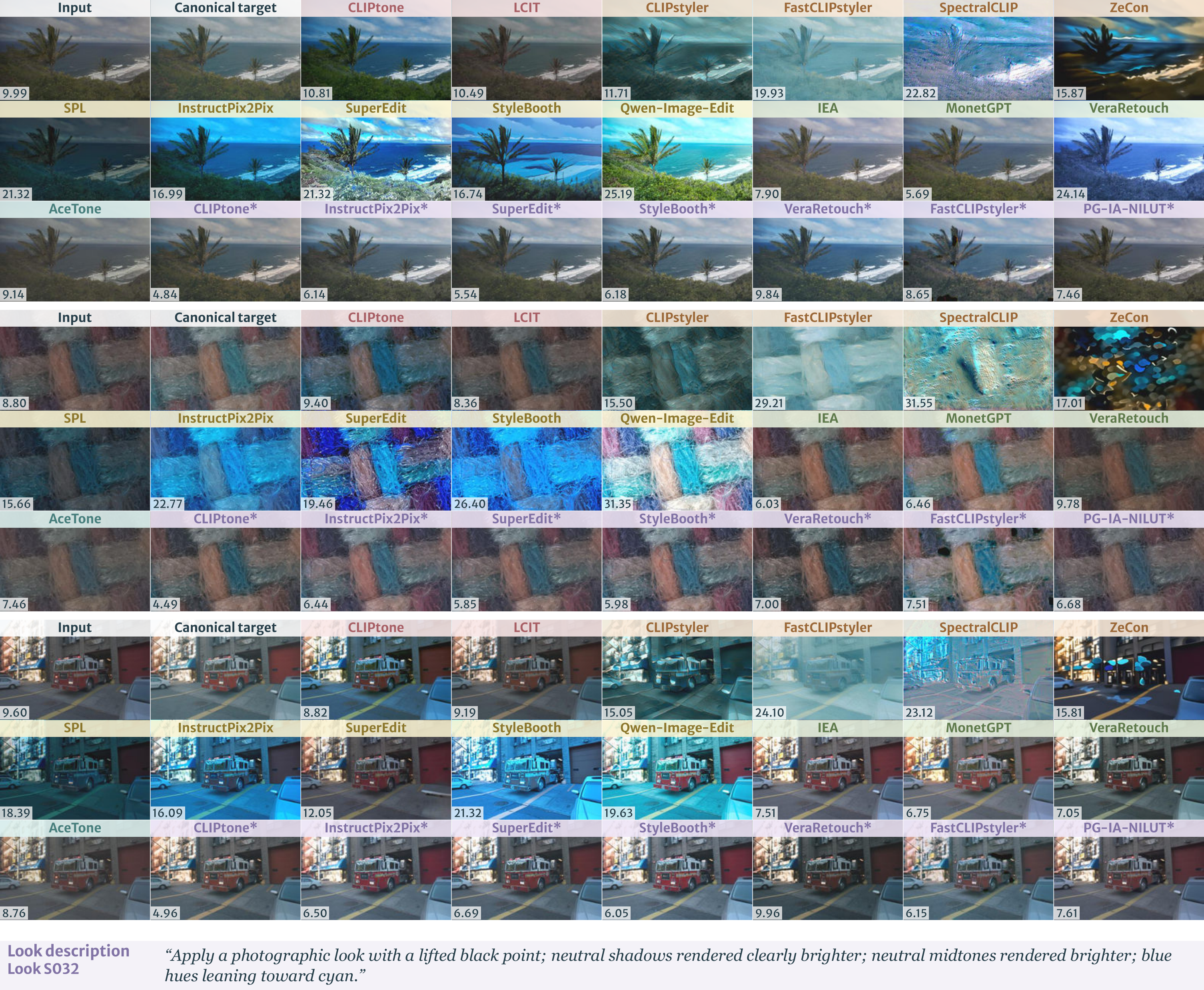}
\caption{Same held-out look and description across three photographs. Each three-row block aligns Input, the canonical target and all 22 method configurations in a fixed order. The shared description is below. Labels give mean pixelwise $\Delta E_{00}\downarrow$ to the canonical target. Input denotes the unedited input. $^*$ denotes LUTBench supervision.}
\label{fig:qualitative}
\end{figure*}

\begin{figure*}[t]
\centering
\includegraphics[width=\textwidth]{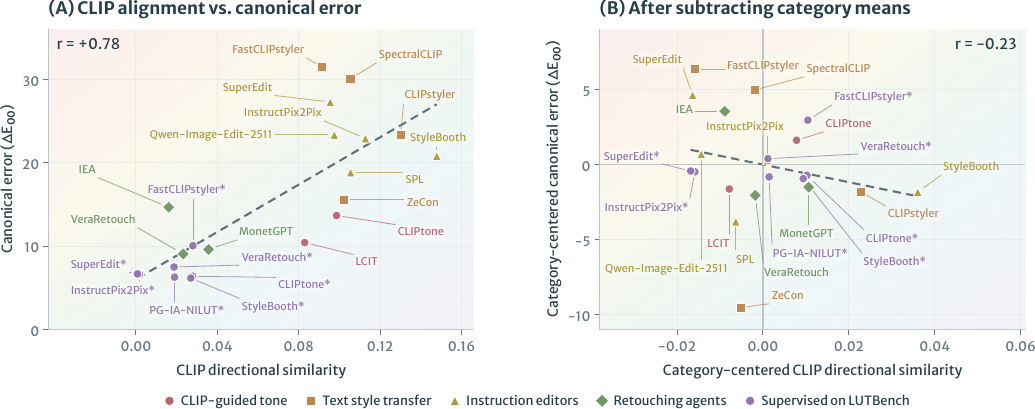}
\caption{Panel A plots raw CLIP directional similarity against canonical error. Panel B subtracts category means from both metrics. Both show 21 configurations. Dashed lines show linear fits, through the origin in B. $r$ is Pearson correlation.}
\label{fig:semantic_fidelity}
\end{figure*}

\begin{figure}[t]
\centering
\includegraphics[width=\columnwidth]{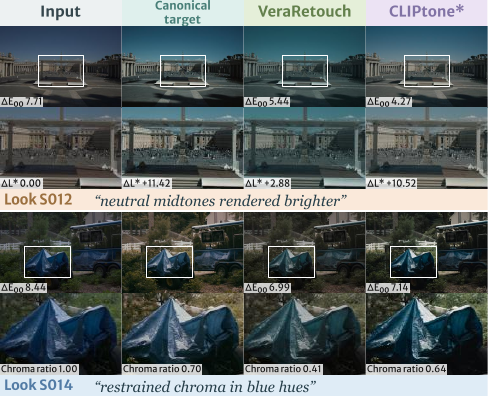}
\caption{Clause satisfaction versus target fidelity. Lightness changes $\Delta L^*$ and output/input chroma ratios are medians over neutral midtone and blue pixels identified in the full input image, respectively. Quoted clauses require $\Delta L^*\geq1.5$ and ratio $\leq0.8$. Boxes mark enlarged regions.}
\label{fig:clause_fidelity}
\end{figure}

\subsection{Target Fidelity and Conversion Sensitivity}
\label{sec:target_fidelity}
We measure canonical error using CIEDE2000. Let $\mathcal I$ denote the evaluated photographs and $\mathcal P_i$ the pixel positions in photograph $i$. For look $l$, canonical error is:
\begin{equation}
 E_l(f)=\frac{1}{|\mathcal I|}
 \sum_{i\in\mathcal I}\frac{1}{|\mathcal P_i|}\sum_{p\in\mathcal P_i}
 \Delta E_{00}(\hat y_{ilp},y_{ilp}),
 \label{eq:error}
\end{equation}
where $\hat y_{ilp}$ and $y_{ilp}$ are the output and canonical target RGB values at pixel $p$, converted to CIELAB for CIEDE2000~\cite{sharma2005ciede2000}. Each look in the evaluation set $\mathcal L$ contributes equally to the aggregate:
\begin{equation}
 \bar E(f)=\frac{1}{|\mathcal L|}\sum_{l\in\mathcal L}E_l(f).
 \label{eq:aggregate}
\end{equation}
To examine sensitivity to input conversion, we keep outputs fixed and vary the target images. Let $E_l(f;r)$ denote the same mean error using targets rendered through conversion $r$. We compute the minimum error over candidate input conversions, abbreviated as minimum error, using the canonical and predefined alternative conversions in $\mathcal R_l$:
\begin{equation}
 E_l^{\min}(f)=\min_{r\in\mathcal R_l}E_l(f;r).
 \label{eq:alternative}
\end{equation}
The evaluator selects one conversion per method and look, shared across all evaluation photographs.

Peak signal-to-noise ratio (PSNR), structural similarity (SSIM)~\cite{wang2004ssim}, and learned perceptual image patch similarity (LPIPS)~\cite{zhang2018lpips} measure pixel, structural, and perceptual fidelity to canonical target images. \Cref*{sec:supp_eval} defines neutral-pixel error, chromatic-pixel error, and hue-balanced color error for locating color deviations. We also report mean absolute weighted CIEDE2000 components to distinguish lightness, chroma, and hue deviations. Signed lightness and chroma biases indicate whether outputs are lighter or darker, or more or less chromatic, than canonical targets. CLIP directional similarity is the cosine similarity between input-to-output image changes and the text direction from the source prompt \textit{``Normal photo.''} to the look description in CLIP embedding space. Clause satisfaction measures compliance with the description for an image pair: it is the weighted fraction of satisfied clauses among those that can be checked using image measurements. Component definitions and both metrics are detailed in \cref*{sec:supp_text_alignment}.

\subsection{Paired Comparisons and Interpretation}
\label{sec:paired_comparisons}
The unedited input serves as a baseline for assessing whether editing improves target fidelity. For each look, we compute the evaluated method's error minus the error of the unedited input or another method configuration. For canonical and minimum errors separately, we estimate a 95\% percentile bootstrap confidence interval for the mean paired difference by resampling looks. An improvement is supported only when both intervals lie below zero. A deficit requires both intervals to lie above zero. Otherwise, the comparison is inconclusive. Agreement between comparisons using canonical and minimum errors shows that conclusions persist when predefined input-encoding alternatives are allowed. We test sensitivity by resampling whole look groups and evaluating look subsets selected by evidence for LUT input encoding or conversion sensitivity, as detailed in \cref*{sec:supp_eval}.

\section{Experiments and Discussion}
\label{sec:results}
\noindent\textbf{Experimental Settings.}
We evaluate 16 methods from prior research in 22 configurations: 14 zero-shot text-driven, one reference-guided, and seven supervised on LUTBench. Zero-shot denotes no training on LUTBench input-target pairs and allows per-image optimization. The zero-shot text-driven configurations span CLIP-guided tone, text style transfer, instruction editors, and retouching agents. Each text-driven method receives the look description formatted as a complete instruction or a style phrase, according to its input interface. AceTone receives a reference pair instead of the description: a training photograph and its canonical target image for the requested look. The supervised variants learn from LUTBench input-target pairs. \Cref*{sec:supp_methods} details the method and test settings.

\noindent\textbf{Zero-Shot and Reference-Guided Transfer.}
Among zero-shot text-driven methods, VeraRetouch has the lowest canonical error in CV, as shown in \cref{tab:main}. Detailed uncertainty estimates are in \cref*{sec:supp_tables}. The unedited input itself has SSIM 0.873 and a canonical error of 8.65 \dE, showing that high structural similarity can coexist with tone and color errors. Reference-guided AceTone improves on the unedited input in CV. Its training reference pair directly specifies input-to-target color changes.

\noindent\textbf{Supervised Transfer Across Both Tracks.}
Paired target supervision reduces canonical error relative to the corresponding zero-shot configurations. For example, InstructPix2Pix improves from 22.20 to 6.51 \dE{} in CV. For VeraRetouch, supervision improves fidelity while keeping the released model's retouching plans fixed and training only the components that turn those plans into image adjustments. This shows that fidelity can improve at the image-adjustment stage without changing the textual plan. The gains vary across architectures. Six of seven supervised variants outperform the unedited input in CV, Test and the Seen track, supporting both reproduction of learned looks and transfer to unseen transformations. \Cref*{sec:supp_seen} presents results for the Seen track and discusses the performance gap with Test.

\noindent\textbf{One Description Across Photographs.}
In \cref{fig:qualitative}, several zero-shot instruction editors overemphasize the description's blue and cyan color cues across the three photographs under Look S032, exceeding the target adjustments. Some outputs also alter local textures. For example, StyleBooth renders coastal cloud and wave detail as broad, flat color regions. Repeating a color bias across photographs does not reproduce the requested transformation. Additional examples appear in \cref*{sec:supp_gallery}.

\noindent\textbf{Semantic Alignment and Target Fidelity.}
Most zero-shot text-driven configurations have higher CLIP directional similarity than the canonical target images in \cref{tab:main}. However, their outputs deviate from the target colors. Across 21 configurations excluding AceTone, higher CLIP directional similarity accompanies larger canonical error in \cref{fig:semantic_fidelity}(A) ($r=0.78$). After subtracting each metric's category mean, the relationship is weakly negative in \cref{fig:semantic_fidelity}(B) ($r=-0.23$). The change after category centering indicates that differences between method categories contribute to the overall association. Cosine similarity is unchanged by positively rescaling an embedding difference vector. Thus, directional alignment does not directly constrain the magnitude of color changes. In a CLIPtone comparison with the pretrained enhancer fixed and only the text adapter trained, target supervision reduces canonical error relative to the released model. In contrast, CLIP supervision increases both directional similarity and canonical error. This comparison shows that stronger semantic alignment does not necessarily improve target fidelity, as detailed in \cref*{sec:supp_objectives}.

\noindent\textbf{Clause Satisfaction and Color Error.}
Clause satisfaction checks the conditions specified by a description, including one-sided bounds on adjustments. Canonical error measures fidelity to the target image. Once a one-sided threshold is met, a stronger adjustment can still satisfy the clause while moving farther from the canonical target. Qwen-Image-Edit-2511 satisfies many measurable clauses. However, its canonical error is large. In \cref{fig:clause_fidelity}, VeraRetouch passes both quoted clause thresholds. However, it brightens neutral midtones less and suppresses blue chroma more than the canonical targets. The supervised CLIPtone variant gives closer values for these two attributes. Lightness and chroma biases in \cref{tab:main} locate deviations from the canonical targets, with definitions in \cref*{sec:supp_text_alignment}. FastCLIPstyler, SpectralCLIP and Qwen-Image-Edit-2511 show the largest positive lightness biases, and CLIPtone and several instruction editors produce higher chroma than their targets. Supervised variants have smaller overall lightness and chroma deviations from their targets, with darker, lower-chroma outputs.

\noindent\textbf{Look Specification and Conversion Sensitivity.}
Look descriptions summarize tone and color adjustments. The LUT and canonical input conversion specify the transformation used to render target images. Identical clause sets can describe different transformations because they summarize adjustment ranges, as detailed in \cref*{sec:supp_descriptions}. LUTBench fixes the canonical input conversion for each look and tests sensitivity to predefined alternatives. Canonical and minimum errors support the same paired-comparison conclusions in CV. Across the input-conversion subsets in \cref*{sec:supp_tables}, all six supervised configurations with supported CV improvements outperform the unedited input.

\noindent\textbf{Limitations and Future Works.}
LUTBench covers global color transformations defined by 3D LUTs, not spatially local edits. Descriptions use a controlled clause vocabulary, which makes each clause verifiable but does not cover free-form requests. In future work, we will reformulate these descriptions as free-form requests that more closely resemble real user prompts and evaluate methods under this setting. We will also extend the evaluation to iterative editing, where users provide feedback on previous outputs to guide subsequent adjustments.

\section{Conclusion}
\label{sec:conclusion}
We introduce LUTBench, a benchmark that connects text descriptions, explicit color transformations and reproducible target images for photographic look transfer. From 91 collected 3D LUTs with canonical input conversions, we render 455,000 input-target pairs across 5,000 photographs. Measurement-based descriptions express each look's tone and color changes in shared terms across photographs. This construction enables direct comparison of how faithfully different methods reproduce the same transformation as image content and initial colors change. The Seen track and Unseen track distinguish reproduction of learned looks from generalization to held-out transformations, with paired comparisons testing sensitivity to input conversion. The evaluation distinguishes semantic alignment from target fidelity. By coupling measurable descriptions with paired targets, LUTBench supports both controlled evaluation and supervised learning. It provides a basis for studying how models translate language into precise color adjustments and for evaluating consistent, controllable look transfer.

{\small
\bibliographystyle{ieeenat_fullname}
\bibliography{references}
}

\appendix
\crefalias{section}{appendix}
\crefalias{subsection}{appendix}
\renewcommand{\thesubsection}{\thesection\arabic{subsection}}
\setcounter{figure}{0}
\setcounter{table}{0}
\renewcommand{\thefigure}{S\arabic{figure}}
\renewcommand{\thetable}{S\arabic{table}}
\renewcommand{\theHfigure}{S\arabic{figure}}
\renewcommand{\theHtable}{S\arabic{table}}
\makeatletter
\twocolumn[{
\centering
{\Large\bfseries \paperfulltitle\par}
\vspace{0.5em}
{\large\bfseries Appendix\par}
\vspace{1.5em}
}]
\makeatother

\section{Dataset and Evaluation Details}
\label{sec:supp_construction}

\subsection{Photograph Preparation and Target Rendering}
\label{sec:supp_photographs}
The source photographs are the 5,000 RAW images of MIT-Adobe FiveK~\cite{bychkovsky2011fivek}. Each RAW image is developed with camera white balance, linear sRGB output, no automatic brightening, half-size demosaicing, 16-bit output and LibRaw's highlight blending (mode 2). A brightness factor of $1/16$ reserves highlight headroom. The decoded values are rescaled to $[0,16]$, resized in linear light to 384 pixels on the long side, and compressed by a bounded highlight function. For a linear channel value $z$, the function is:
\begin{equation}
 h(z)=\begin{cases}
 z,&z\leq 0.75,\\
 0.75+a\log\!\left(1+(z-0.75)/a\right),&z>0.75,
 \end{cases}
\end{equation}
where $a\approx0.04247$ makes $h(16)=1$. The compressed channel values are sRGB-encoded and saved as a 16-bit PNG. The saved image defines the input photograph $x_i$ for editing and for rendering canonical and alternative target images. \Cref*{sec:supp_conversions} details conversion and interpolation.

We partition the 5,000 photographs into 4,000 training, 500 validation and 500 test photographs.

\subsection{Input Conversions and Their Alternatives}
\label{sec:supp_conversions}
The 91 LUTs comprise 87 grids of size $33^3$ and four of size $65^3$. The canonical input conversion is chosen using input-encoding metadata or calibration against matched LUT exports.

For a logarithmic LUT, the saved photograph is first decoded with the inverse sRGB transfer function and then mapped through $h^{-1}$ to recover linear values before highlight compression. The conversion multiplies these values by $2^{\Delta\mathrm{EV}}$, applies the matrix from linear sRGB to the selected RGB primaries, and encodes the result with the selected logarithmic curve. For the canonical display-input conversion, the saved sRGB values enter the LUT directly. Before trilinear interpolation, LUT inputs are clipped to the domain declared in the LUT file. LUT outputs are clipped to $[0,1]$ and interpreted as display sRGB.

Input-conversion calibration compares logarithmic and sRGB exports matched by collection label and normalized LUT name, using the sRGB export as a reference. These matched export sets are evaluated on the same training photographs. The exposure search covers $-0.5$ to $2.5$ EV in $0.25$-EV increments. S-Log3 candidates use sRGB, S-Gamut3.Cine or S-Gamut3 primaries. C-Log and C-Log3 candidates use sRGB or Cinema Gamut. CIEDE2000 differences are averaged over sampled pixels, photographs and export pairs within each matched export set. The calibration distance is the median of these set averages. The minimum-distance settings use S-Gamut3.Cine at $+1.75$ EV for S-Log3 and Cinema Gamut at $+1.50$ EV for both C-Log and C-Log3. For the uncalibrated logarithmic conversions, encoding names come from LUT filenames or associated metadata. The fixed settings use BT.2020 at zero EV for Apple Log, V-Gamut at $+1$ EV for V-Log, and BT.2020 at $+1$ EV for F-Log.

Alternative input conversions test sensitivity to the selected settings. For calibrated logarithmic inputs, alternatives use exposure-offset and RGB-primary settings whose calibration distance is at most 1.10 times the minimum over tested settings. For uncalibrated logarithmic inputs, the alternatives vary the exposure offset by $\pm0.5$ EV and include sRGB primaries. The display-input alternative replaces sRGB code values with BT.1886 gamma-2.4 values by applying sRGB decoding followed by a $1/2.4$ power. For the canonical Apple Log inputs of looks S054 and S089, the alternative set additionally includes S-Log3. For the canonical C-Log input of look S064, the alternative set additionally includes C-Log3.

For sensitivity analysis, we distinguish four input-conversion categories: 45 looks with calibrated logarithmic conversions, 28 with display-encoded inputs, 15 with uncalibrated logarithmic conversions, and three with alternatives across logarithmic encodings. For each look, we compute the mean CIEDE2000 difference between its canonical and alternative target images on the 500 validation photographs, separately for each alternative conversion. Conversion sensitivity is the largest of these means. It is at least 4 $\Delta E_{00}$ for 25 training looks and 5 validation looks, and below 4 for all test looks. \Cref{tab:supp_reliability} summarizes sensitivity by input-conversion category.

\begin{table}[t]
\centering\footnotesize
\caption{Sensitivity to the LUT input conversion, measured by the difference between canonical and alternative target images on 500 validation photographs. The last column gives the median, across looks, of the largest mean $\Delta E_{00}$ over alternatives.}
\label{tab:supp_reliability}
\begin{tabular}{@{}lcc@{}}
\toprule
Conversion category & Looks & Median largest shift \\
\midrule
Calibrated log & 45 & 2.59 \\
Display & 28 & 3.87 \\
Uncalibrated log & 15 & 4.59 \\
Cross-encoding alternatives & 3 & 7.43 \\
\bottomrule
\end{tabular}
\end{table}

\subsection{Color Measurements and Look Descriptions}
\label{sec:supp_descriptions}
We construct 497 fixed sRGB color samples. The 17 neutral grays have $R=G=B=k/16$ for $k=0,\ldots,16$, including black and white. Chromatic samples use all combinations of six fixed HSV hues at $0^\circ$, $60^\circ$, $120^\circ$, $180^\circ$, $240^\circ$ and $300^\circ$ (red, yellow, green, cyan, blue and magenta), five saturation values $S\in\{0.15,0.30,0.50,0.75,1.00\}$, and 16 equally spaced value levels from $V=0.1$ to $V=1.0$, including both endpoints. Converting these HSV triples to sRGB gives $6\times5\times16=480$ chromatic samples. Each look's LUT is applied to this common sample set under its canonical input-conversion settings.

Fixed color samples are specified directly in sRGB and use standard sRGB decoding without $h^{-1}$ before logarithmic input conversion. In contrast, the saved-photograph path in \cref*{sec:supp_conversions} also reverses the highlight compression defined in \cref*{sec:supp_photographs}. Therefore, identical sRGB highlight values can map to different LUT inputs in the two paths. White-point clauses describe the fixed white sample.

The measurements include the black and white points, contrast along the neutral tone curve, the CIELAB coordinates of neutral samples in shadows, midtones, and highlights, and the median hue rotation and chroma ratio for each fixed hue. Let $L^{\mathrm{in}}(k)$ and $L_l^{\mathrm{out}}(k)$ be the input and output lightness of neutral sample $k$ under look $l$. Contrast is the ratio of output to input lightness differences between the gray levels $3/16$ and $12/16$:
\begin{equation}
 \kappa_l=\frac{L_l^{\mathrm{out}}(12)-L_l^{\mathrm{out}}(3)}{L^{\mathrm{in}}(12)-L^{\mathrm{in}}(3)}.
 \label{eq:supp_contrast}
\end{equation}
Neutral shadows, midtones and highlights are defined by the input sample's lightness $L^*$ in the intervals $[5,35)$, $[35,70)$ and $[70,101)$. The upper bound 101 includes white at $L^*=100$. Tone descriptions use the LUT-rendered sample's lightness minus the input sample's lightness. Chroma ratios use samples with input $C^*>12$. Hue rotations additionally require output $C^*>5$ and use output-minus-input CIELAB hue angles, wrapped to $[-180^\circ,180^\circ)$.

\Cref*{fig:look_diversity} in the main paper summarizes the 91 looks on the same 497 fixed color samples. Across all 4,095 unordered look pairs, mean CIEDE2000 differences have a median of 11.61, with the central half between 9.46 and 14.16 and a range from 0.64 to 31.13. Chroma reduction varies by hue: median output-to-input chroma ratios are 0.50 for blue and 0.84 for red.

For \cref*{fig:look_color_space} in the main paper, each look is represented by the $(\Delta L^*,\Delta a^*,\Delta b^*)$ changes over the 497 fixed samples, giving $497\times3=1{,}491$ coordinates. Panel A averages the $a^*$ and $b^*$ changes over samples. For panel B, we divide each complete vector by $\sqrt{497}$ and center each coordinate across the 91 looks without variance standardization before fitting PCA. Every sample and CIELAB coordinate has equal weight. The background in B is a Gaussian-weighted local mean of lightness changes, with bandwidth 7\% of each axis span and opacity fading away from measured points. Labels identify coordinate extrema.

\noindent\textbf{Clause Construction and Validation.}
\Cref{tab:supp_vocabulary} defines the eight clause templates. For color cast, chroma and regional lightness, we also define clauses that specify adjustment direction with less restrictive numerical conditions. A clause describing a lightness change in neutral shadows, midtones, or highlights uses the median difference in $L^*$ between the canonical target image and input photograph. The measurement includes input pixels in the corresponding lightness interval with CIELAB chroma $C^*<8$. Color-cast measurements use the median target $a^*$ or $b^*$ at these same initially neutral pixels. The templates name positive and negative $a^*$ as magenta and green casts, respectively, and positive and negative $b^*$ as yellow and blue casts.

For photograph measurements, HSV hue $H$ in degrees assigns pixels to six centered hue sectors using $j=\lfloor(H+30^\circ)/60^\circ\rfloor\bmod6$, in the order red, yellow, green, cyan, blue and magenta. The red sector spans the $360^\circ/0^\circ$ boundary. Chroma ratios and hue rotations use input pixels with $C^*>12$ and $5<L^*<95$. Hue rotations also require target $C^*>5$. Each photograph-level attribute is measurable only when at least 16 eligible pixels are present. The ratio and angular measurements otherwise follow the fixed-sample definitions above.

For each attribute, an accepted clause specifying adjustment direction is used only when no clause specifying adjustment strength passes validation. Clauses specifying tone or color-cast direction require the stated sign and a magnitude of at least 1.5 in the relevant CIELAB coordinate. Clauses specifying chroma direction use ratios at most 0.9 or at least 1.1 for reduction or enhancement. Strength labels use non-overlapping bins. For tone and color-cast magnitudes, these are $[2,5)$, $[5,12)$ and $[12,\infty)$. In contrast, clause satisfaction checks include both finite endpoints of the intervals in \cref{tab:supp_vocabulary}.

To order the accepted clauses, let $m_{lj}$ be the fixed-sample measurement of attribute $j$ for look $l$, $\nu_j$ its reference value for no adjustment, and $\sigma_j$ its population standard deviation over available measurements across the 91 looks. For clause $c$ describing attribute $j(c)$, the ranking score is:
\begin{equation}
 q_{lc}=\frac{|m_{l,j(c)}-\nu_{j(c)}|}{\max(\sigma_{j(c)},10^{-6})}\,\widetilde{s}_{lc}.
 \label{eq:supp_clause_ranking}
\end{equation}
Here, $\widetilde{s}_{lc}=s_{lc}$ for clauses checked using image measurements and $0.8$ for clauses checked only on fixed color samples. The reference values for no adjustment are one for chroma and contrast ratios, 100 for white-point lightness, and zero for the remaining attributes. The description includes up to two tone clauses, two chroma clauses, one color-cast clause, and one hue clause before filling the remaining places in score order. An initial allocation uses only the accepted clauses available in its attribute group. The maxima are four tone, four chroma, three color-cast and three hue clauses. At most two clauses describe regional lightness, and each description contains at most eight clauses. Each description contains 3--8 clauses and 19--54 words and fits within CLIP's 77-token input limit. The library contains 651 occurrences of 73 distinct clauses.

We compare five description variants using LUT predictors conditioned on binary indicators of clause presence, bag-of-words features, or frozen CLIP text embeddings. The selected rule gives the lowest mean CIEDE2000 error across these three models on 40 held-out CV looks and validation photographs. Test-look descriptions use the same fixed-color measurements and training-photograph checks. Attribute normalization and inverse-frequency clause weights are computed across the 91-look library and shared by all methods. No method is trained on LUTBench test-look image pairs.

On validation photographs, the 541 clause occurrences checked using image measurements achieve 93.1\% support for their full numerical conditions and 98.8\% agreement in direction alone. Directional agreement checks the sign of lightness changes, hue rotations and color-cast coordinates, and whether chroma ratios are above or below one, as specified by the clause. Both rates use the inverse-frequency weights in \cref*{sec:supp_text_alignment} to average measurable clauses within each photograph, then over photographs and equally over all 91 looks, including the 10 test looks.

\noindent\textbf{Description Similarity and Look Distance.}
Description overlap is the Jaccard similarity between clause sets. Look distance is the mean CIEDE2000 difference between corresponding canonical target images on validation photographs. Across 3,240 pairs of the 81 non-test looks, the Spearman correlation between these measures is $-0.49$, associating greater description overlap with more similar rendered effects. Two pairs of training looks share clause sets. For one pair, the descriptions are identical and the look distance is 0.41. For the other pair, the descriptions differ only in clause order and the look distance is 3.66.

\begin{table*}[t]
\centering
\footnotesize
\setlength{\tabcolsep}{4pt}
\caption{Candidate-clause templates and numerical conditions. Uses counts clause occurrences across the 91 descriptions. Conditions marked fixed samples use fixed color samples, and the other conditions use training photographs. The lightness change $\Delta L^*$ is measured from input to canonical target image, and color casts use $a^*$ and $b^*$ in the canonical target image at initially neutral pixels. Within each hue sector, the chroma ratio is the median ratio of CIELAB chroma in the canonical target image to that in the input photograph.}
\label{tab:supp_vocabulary}
\begin{tabular}{@{}p{1.15cm}p{7.05cm}p{7.0cm}c@{}}
\toprule
Group & Clause & Holds when & Uses \\
\midrule
Tone & \textit{``neutral \{zone\} rendered \{slightly, clearly, much\} \{brighter, darker\}''} & magnitude of median $\Delta L^*$ for neutral pixels in $[2,5]$, $[5,12]$, or $[12,\infty)$, with sign matching brighter/darker. Directional variant: $\geq1.5$ & 129 \\
 & \textit{``a lowered white point''} & white sample lightness $L^* \leq 95$ (fixed samples) & 49 \\
 & \textit{``a lifted black point''} & black sample lightness $L^* \geq 4$ (fixed samples) & 25 \\
 & \textit{``\{soft, pronounced\} contrast along the neutral tone curve''} & contrast ratio $\leq 0.85$ or $\geq 1.15$ (fixed samples) & 36 \\
Color cast & \textit{``neutral \{zone\} rendered with a [\{subtle, clear, strong\}] \{color\} cast''} & magnitude of the zone-median $a^*$ or $b^*$ in $[2,5]$, $[5,12]$, or $[12,\infty)$, with sign matching the color cast. Directional variant: $\geq1.5$ & 138 \\
Chroma & \textit{``\{restrained, slightly restrained\} chroma in \{hue\} hues''} & chroma ratio $\leq 0.8$, $\leq 0.9$ & 141 \\
 & \textit{``\{rich, slightly enhanced\} chroma in \{hue\} hues''} & chroma ratio $\geq 1.2$, $\geq 1.1$ & 5 \\
Hue & \textit{``\{hue\} hues leaning toward \{neighbor\}''} & hue rotation of 5 to 65$^\circ$ toward the neighbor & 128 \\
\bottomrule
\end{tabular}
\end{table*}

\subsection{Look Groups, Partitions and Evaluation Scope}
\label{sec:supp_splits}
The distance between two looks is the mean CIEDE2000 difference between their canonical target images at corresponding pixel positions. This distance is measured on 512 pixels from each of 16 fixed training photographs. The training photograph identifiers are sorted, shuffled with NumPy PCG64 at seed 2026, and selected from the end of the shuffled list. A separate PCG64 generator at seed 2026 uniformly samples 512 distinct pixels without replacement from each selected photograph in that order. Looks less than 4 $\Delta E_{00}$ apart are linked, and connected components define 42 look groups. There is one group of 28 looks, one of 7, one of 5, three of 3, six of 2, and 30 singletons. The fixed training, validation, and test partitions contain 72/9/10 looks and keep these groups intact, with minimum cross-partition distance 4.37 $\Delta E_{00}$.

\Cref*{fig:look_groups} in the main paper visualizes group sizes and pairwise distances. In the inset, boxes show quartiles and medians, whiskers span the full range, and the dashed line marks the edge threshold of 4 $\Delta E_{00}$. The median distance is 6.06 within groups, 9.47 between groups, and 10.54 across partitions. Each unordered look pair has equal weight. The largest group contributes 378 of the 424 within-group pairs.

The 40 non-test looks form five fixed held-out subsets of eight. For each supervised variant, five models are trained, each excluding a different subset and using the remaining 73 non-test looks. Results are pooled across the five held-out subsets. The 40 CV looks span 29 look groups. CV and Val share 9 evaluation looks, with supervised models trained on different look partitions.

CV and Val use all 500 validation photographs. Test evaluates the 10 test looks and the Seen track evaluates the 72 training looks, both on all 500 test photographs.

\subsection{Metrics, Intervals and Sensitivity Analysis}
\label{sec:supp_eval}
The evaluator uses NumPy, colour-science, OpenCV and Pillow. Output images are resized to the input photograph dimensions with Lanczos interpolation before evaluation. CIEDE2000~\cite{sharma2005ciede2000} uses $k_L=k_C=k_H=1$. Canonical error and minimum error follow the definitions in \cref*{sec:target_fidelity}. The complementary image metrics and sensitivity analyses are selected using validation data.

Neutral-pixel error is computed at input pixels with CIELAB chroma $C^*<8$. Chromatic-pixel error uses input pixels with $C^*>12$ and lightness $5<L^*<95$. Hue-balanced color error groups these chromatic pixels into the six input-HSV hue sectors defined in \cref*{sec:supp_descriptions}, then averages sector errors equally over sectors containing at least 200 pixels. Membership is determined from the input photographs, and pixels are pooled across photographs before computing each error. PSNR is computed for each photograph, then averaged across photographs and looks. RGB SSIM~\cite{wang2004ssim} uses the scikit-image defaults, including a $7\times7$ uniform window and sample covariance. LPIPS~\cite{zhang2018lpips} uses AlexNet. SSIM and LPIPS compare output images with canonical target images at the saved input photograph resolution.

For the paired comparison in \cref*{sec:paired_comparisons}, 95\% percentile bootstrap confidence intervals (CIs) use 2,000 resamples of looks.

Two sensitivity analyses supplement the rule. For look-group resampling, each replicate draws as many groups as the comparison contains, with replacement, and includes every look in each drawn group. Let $\delta_l$ be the paired error difference for look $l$ and $\mathcal G^{(b)}$ the multiset of groups drawn in replicate $b$. Its mean gives each included look equal weight:
\begin{equation}
\bar\delta^{(b)}=
\frac{\sum_{g\in\mathcal G^{(b)}}\sum_{l\in g}\delta_l}
{\sum_{g\in\mathcal G^{(b)}}|g|}.
 \label{eq:supp_group_bootstrap}
\end{equation}
A group drawn more than once contributes once per draw. Percentile bounds are computed from 2,000 replicate means. Three input-conversion subsets contain looks with explicit evidence for LUT input encoding, looks whose largest mean difference between canonical and alternative target images is below 4 $\Delta E_{00}$, and their intersection. The three CV subsets contain 32, 21 and 13 looks, respectively.

\subsection{Color Components and Text Alignment}
\label{sec:supp_text_alignment}
\noindent\textbf{CIEDE2000 Components.}
The weighted CIEDE2000 components $\Delta L'/S_L$, $\Delta C'/S_C$ and $\Delta H'/S_H$ use color differences computed as values in the output image minus values in the canonical target image. The component measurements use the evaluation pixels from the corresponding photograph sets.

\noindent\textbf{Text Alignment.}
CLIP directional similarity follows CLIPtone~\cite{lee2024cliptone} using normalized OpenAI ViT-B/16 embeddings~\cite{radford2021clip}. It is the cosine similarity between the embedding difference from the input photograph to the output image and the text-embedding difference from the source prompt \textit{``Normal photo.''} to the description followed by \textit{``photo.''} The image direction is zero for the unedited input, making its CLIP directional similarity undefined.

Clause satisfaction uses inverse-frequency weights to aggregate the selected clauses that can be checked using image measurements. Let $N=91$ and let $n_c$ count occurrences of clause text $c$ in the candidate pool across these looks, before acceptance and selection, including candidate clauses specifying adjustment strength or direction. The fixed weight uses the natural logarithm:
\begin{equation}
w_c=1+\log\frac{1+N}{1+n_c}.
 \label{eq:supp_clause_weight}
\end{equation}
For input $x_i$ and output $\hat y_{il}$, let $\mathcal A_{il}(f)$ contain the selected clauses whose attributes can be measured in that pair, and let $t_{ilc}(f)$ be one if the output satisfies clause $c$ and zero otherwise. The pair's clause satisfaction is:
\begin{equation}
\operatorname{CS}_{il}(f)=
\frac{\sum_{c\in\mathcal A_{il}(f)}w_c\,t_{ilc}(f)}
{\sum_{c\in\mathcal A_{il}(f)}w_c}.
 \label{eq:supp_clause_satisfaction}
\end{equation}
Pairs with no applicable clauses are omitted from the mean. We average over photographs within each look, then equally over looks. Of the 541 clause occurrences checked using image measurements across the 91 descriptions, 334 impose one-sided thresholds, including every chroma clause. A one-sided condition constrains the adjustment to one side of a threshold without fixing its magnitude.

CLIP directional similarity and clause satisfaction use all 40 CV looks and all 500 validation photographs. Each method follows the inference settings of the target-fidelity evaluation. The CLIPtone models trained independently for the five CV held-out subsets follow the same training procedure as the corresponding canonical-error comparison. The canonical target images provide reference values of 94\% clause satisfaction and 0.036 CLIP directional similarity.

\noindent\textbf{Association with Target Fidelity.}
For \cref*{fig:semantic_fidelity} in the main paper, CLIP directional similarity and canonical error are computed from the same output images. Each 8-bit RGB output is compared with its 16-bit canonical target. The reference-guided category contains only AceTone. Therefore, both panels compare the same 21 configurations from the other five method categories in \cref*{tab:main} in the main paper.

Panel A uses the original configuration means and an ordinary least-squares fit with an intercept. Panel B subtracts each category's configuration mean from both coordinates, without standard-deviation normalization. Its dashed line is a pooled least-squares fit through the origin, equivalent to fitting a common slope with category-specific intercepts.

\section{Experimental Settings}
\label{sec:supp_methods}

\subsection{Common Implementation Settings}
\label{sec:supp_interfaces}
\label{sec:supp_training}
Instruction interfaces receive the complete look description $d_l$. Style-phrase interfaces remove \textit{``Apply a photographic look with''} and the final period. In the prompt templates below, \textit{``\{d\}''} denotes this remaining style phrase. The default random seed is $2026\cdot10^6+1000l+i$, where $l$ is the look index and $i$ is the photograph position in the evaluation set. The default inference platform is Windows with a 32 GB RTX 5090. Diffusion editors process batches of eight photographs with matching dimensions. Learning-rate selection uses the first CV held-out subset, and the selected rate is fixed for the remaining models.

The supervised InstructPix2Pix, SuperEdit and StyleBooth variants share a diffusion training recipe. The U-Net is trained with the variational autoencoder and text encoder frozen. Each uses 600 training photographs per look. Both images in each training pair are quantized to 8-bit RGB and directly resized to $256\times256$, followed by paired horizontal flips. AdamW runs for 15,000 steps at batch size 16 and learning rate $5\cdot10^{-5}$. Training uses bf16 autocast, conditioning dropout 0.05, and mean squared error between predicted and sampled noise. Each variant uses its own instruction template, diffusion scheduler and released inference settings.

Parameter counts cover one instantiated editing system, including frozen encoders and counting shared weights once. Counts use checkpoint tensor shapes and model definitions, excluding fixed renderers, fitted per-image color coefficients and evaluation networks.

\subsection{Evaluated Methods}
\label{sec:supp_method_profiles}
\label{sec:supp_original_data}

\noindent\textbf{CLIPtone (CVPR, 2024)}~\cite{lee2024cliptone}.
The 155.6M-parameter configuration combines an AiLUT enhancer with an RN50 text adapter. It uses the direction from \textit{``Normal photo.''} to the complete description without its final period followed by \textit{``photo.''}, at intensity 1 and the saved image resolution. LUT lookup follows the released kernel's axis convention. On LUTBench, the main supervised variant trains the enhancer and text adapter from scratch using the initialization routines in the released code, with one identity basis LUT and the remaining basis LUTs initialized to 0. Each step uses 64 photographs, four looks and 512 pixels per photograph. Training runs for 8,000 steps with L1 loss, smoothness weight $10^{-3}$ and monotonicity weight $10^{-2}$. AdamW uses $3\cdot10^{-4}$, selected by canonical error from $\{3\cdot10^{-4},10^{-3},3\cdot10^{-3}\}$. The supervised adapter and full fine-tuning variants use the same losses at $3\cdot10^{-4}$ and $3\cdot10^{-3}$. These supervised variants use 200 warm-up steps followed by cosine learning-rate decay. The CLIP-objective adapter variant freezes the enhancer and uses 32 crops per step, Adam at $10^{-3}$, content and directional weights 1, weight regularization $10^{-4}$ and interval regularization 0.5 for 8,000 steps. Original training uses 4,500 FiveK inputs and the Color Names Dataset. Testing uses 500 FiveK photographs and 50 tone descriptions. A reference-based comparison uses 9 Lightroom filters.

\noindent\textbf{LCIT (CIC, 2025)}~\cite{mayer2025lcit}.
The 151.3M-parameter configuration fits a white-preserving $3\times3$ color matrix in linear light using OpenCLIP ViT-B/32 with LAION-2B weights. LCIT is implemented from the \href{https://library.imaging.org/cic/articles/33/1/38}{paper}. The prompt is \textit{``A \{d\} photo''}. Loss computation uses a 224-pixel center crop, matrix offsets are bounded by 0.25, and Adam runs for 1,000 iterations per photograph at $2\cdot10^{-3}$. Original tests use 24 public Kodak Lossless True Color images with color, cultural and emotion prompts.

\noindent\textbf{CLIPstyler (CVPR, 2022)}~\cite{kwon2022clipstyler}.
The 158.3M-parameter configuration uses the fast decoder initialized from AdaIN, with a random seed set per look. One decoder is fitted per look for 200 iterations on DIV2K crops, using the style phrase in 79 ImageNet templates and source prompt \textit{``a Photo''}. Inference preserves the full frame at long side 512, with contrast 1.5 and min-max stretching. In contrast, the original per-image variant optimizes on input crops. Original demonstrations include content examples and CelebA-HQ facial style transfer.

\noindent\textbf{FastCLIPstyler (WACV, 2024)}~\cite{suresh2024fastclipstyler}.
The 153.5M-parameter configuration encodes the style phrase with CLIP ViT-B/32 and applies the \href{https://github.com/sertiscorp/FastCLIPstyler}{released stylization network} deterministically at the saved image size. The LUTBench supervised variant freezes CLIP and trains the text embedder and stylization network with L1 loss on 200 photographs per look, random $208^2$ crops and flips. AdamW runs for 8,000 steps at batch size 16 and rate $3\cdot10^{-3}$, selected by canonical error from $\{3\cdot10^{-4},10^{-3},3\cdot10^{-3}\}$. Training uses 200 warm-up steps followed by cosine learning-rate decay. Original training uses 4,300 keyword-composed prompts and 1,500 ChatGPT-generated prompts, with a stylizer pretrained on Painter by Numbers and DTD. Testing uses 80 prompts, partly drawn from DTD and ArtEmis, and AdaAttn-rendered semi-reference targets.

\noindent\textbf{SpectralCLIP (WACV, 2024)}~\cite{xu2024spectralclip}.
The 171.9M-parameter configuration introduces spectral filtering into CLIPstyler's per-image optimization. It uses the same phrase templates, long side 512, 200 optimization steps with 64 crops, spectral band c2, contrast 1.5 and min-max stretching. Up to four independent image-specific networks are optimized in parallel. The original evaluation comprises 100 image-style groups using COCO validation images.

\noindent\textbf{ZeCon (ICCV, 2023)}~\cite{yang2023zecon}.
The 722.5M-parameter configuration uses the public unconditional ADM backbone at $256^2$ resolution. It receives the style phrase as the target prompt and \textit{``Photo''} as the source prompt. Inference uses DDIM forward and DDPM reverse sampling, with a 50-timestep schedule, 25 skipped timesteps and one output sample per input. The input is resized to $256\times256$, and the output is restored to the saved input dimensions. Original tests use FFHQ, CelebA-HQ, ImageNet, LSUN-Church and CycleGAN datasets, plus WikiArt for unseen-domain transfer.

\noindent\textbf{SPL (WACV, 2026)}~\cite{gong2026spl}.
The 1.066B-parameter configuration performs training-free optimization with LCM-Dreamshaper-v7. It uses the text direction from \textit{``a photo''} to \textit{``a photo with \{d\}''}, 15 sampling steps, source and target guidance scales of 1 and 2, respectively, and 100 optimization iterations. Inputs are resized to $512\times512$, with outputs restored to the saved dimensions. Original tests use curated PIE-Bench and AnyEdit subsets, with FiveK for tone adjustment and DPST for photographic style transfer.

\noindent\textbf{InstructPix2Pix (CVPR, 2023)}~\cite{brooks2023instructpix2pix}.
The 1.066B-parameter editor receives the full description and processes photographs at long side 512. Released inference uses 100 steps, text guidance 7.5 and image guidance 1.5. The LUTBench supervised variant updates the U-Net using the shared diffusion recipe in \cref*{sec:supp_training}. Original training constructs about 450,000 edits from LAION-Aesthetics V2 6.5+ captions, GPT-3 instructions and Prompt-to-Prompt pairs. Evaluation compares CLIP similarities across 2,000 edits and presents real-photo and artwork examples.

\noindent\textbf{SuperEdit (ICCV, 2025)}~\cite{li2025superedit}.
The 1.066B-parameter SD1.5 configuration uses the full description, long side 512, 50 Euler ancestral steps, text guidance 10, image guidance 1.5 and fp16 inference. The LUTBench supervised variant trains the U-Net with the shared diffusion recipe in \cref*{sec:supp_training}, using the same inference settings. The original 40,000 training pairs combine InstructPix2Pix (10,177), MagicBrush (8,807) and Seed-Data-Edit (21,016). Original tests use the public MultiReward Real-Edit and MagicBrush single-turn test sets.

\noindent\textbf{StyleBooth (ICCV Workshops, 2025)}~\cite{han2024stylebooth}.
The 1.066B-parameter text checkpoint uses SCEPTER with prompt \textit{``Let this image be in the style of \{d\}''}. Full-frame inference at long side 1024 uses 50 DDIM steps, text guidance 7.5, image guidance 1.5, rescale 0.5 and fp16. PyTorch fused attention preserves the released scale. The supervised variant uses the shared diffusion recipe in \cref*{sec:supp_training} with StyleBooth's template and scheduler. Original training uses the StyleBooth Dataset, generated with SDXL and Fooocus templates. Original tests use 226 public style edits from \href{https://huggingface.co/datasets/facebook/emu_edit_test_set}{Emu Edit}.

\noindent\textbf{Qwen-Image-Edit-2511 (Model Release, 2025)}~\cite{wu2025qwenimage}.
The 28.85B-parameter configuration uses the full description, native pipeline resolution, 40 steps, true CFG 4.0 and a blank negative prompt. Inference uses bf16 without weight quantization on H100 80 GB GPUs, with one model instance per GPU. The cited series report describes collected image-text data and synthetic edits, and evaluates on the public GEdit-Bench and ImgEdit benchmarks.

\noindent\textbf{IEA (CVPR Findings, 2026)}~\cite{iea2026}.
The 8.292B-parameter configuration uses a Qwen2.5-VL-7B policy to select among 16 global editing tools. IEA receives the full description and the input photograph at its saved resolution. It uses greedy decoding, up to 1,024 tokens and three attempts. Inference uses the released messages and decoding settings through Transformers. Original training distills tool trajectories from GIER and synthesizes editing and refinement examples from GIER and FiveK. Original tests use these public datasets and a user study of 50 GIER examples.

\noindent\textbf{MonetGPT (TOG, 2025)}~\cite{dutt2025monetgpt}.
The 8.291B-parameter Qwen2-VL-7B configuration appends the full description to each of three editing stages, using style \textit{``balanced''}, temperature 0.3, top-$p$ 0.7 and top-$k$ 50. It uses the released messages and decoding settings through Transformers, with the random seed set per batch of 25 photographs and NumPy and GIMP 2.10 rendering at the saved resolution. Original training constructs visual puzzles from PPR10K Expert A retouches. Cross-dataset testing uses 400 FiveK photographs, and user studies include FiveK and Reddit images.

\noindent\textbf{VeraRetouch (SIGGRAPH, 2026)}~\cite{veraretouch2026}.
The 634.0M-parameter configuration combines FastVLM-0.5B with a renderer. Its retouch head maps VLM hidden states to parameters applied by the renderer. The full description enters the style-mode template, with VLM short side 512, temperature 0.2, top-$p$ 0.8 and one attempt. Rendering uses the saved image resolution. The LUTBench variant freezes the VLM and trains the retouch head and renderer with L1 loss on 16,384 sampled pixels per output. For each training look, the released model generates a retouching plan for each of 16 training photographs, and the corresponding VLM hidden states provide the retouch head inputs. AdamW uses batch size 16, 4,000 steps and rate $10^{-4}$, selected by validation L1 loss from $\{3\cdot10^{-5},10^{-4},3\cdot10^{-4}\}$. Training uses 200 warm-up steps followed by cosine learning-rate decay. At evaluation, the frozen VLM receives the current photograph, its look description, and the zero-shot retouching plan previously generated for that photograph and look. The plan is supplied as fixed text, and the resulting hidden states feed the trained retouch head. Original training uses AetherRetouch-1M+, combining FiveK and PPR10K expert retouches with Unsplash-Full images and 5,030 Lightroom presets. Tests cover Aether-Bench, FiveK-Bench and supplementary PPR10K-Bench results.

\noindent\textbf{AceTone (CVPR, 2026)}~\cite{ma2026acetone}.
The evaluated 3.760B-parameter PST preview uses a reference pair instead of a text description: one training photograph and its canonical target image under the requested look. For each look, the reference photograph is drawn uniformly from the training photographs using a fixed seed specific to that look. The same reference pair is used for all evaluated input photographs under that look. Greedy decoding predicts 64 LUT tokens. The decoded LUT is applied at the saved image resolution using the released channel and lattice conventions. Original training uses MS-COCO pretraining and FiveK/PPR10K fine-tuning, including about 800,000 instruction-guided grading examples. Testing uses AceTone-Bench Instruct (128 examples), Transfer (1,024) and PST-50.

\noindent\textbf{PG-IA-NILUT (ACM MM, 2024)}~\cite{pgianilut2024}.
The 0.449M-parameter FiveK model converts five sliders into one $33^3$ LUT, applied at the saved resolution with trilinear interpolation at the released implementation's uniform grid vertices. For LUTBench, five slider values are fitted per training look using L1 loss, 256 photographs and 512 pixels per photograph at each step. Adam runs for 400 steps at 0.05 with cosine decay and slider bounds $[-1,1]$. Ridge regression predicts slider values from a binary vector indicating which description clauses are present. The vocabulary contains the $K$ distinct clauses in the current training-look descriptions, with one additional entry for clauses outside this vocabulary, giving $K+1$ features. Each slider is standardized using its mean and standard deviation over the training looks, with the standard deviation bounded below by $10^{-6}$. The regression uses an unpenalized intercept and selects among 13 penalties, $\{10^{-3},10^{-2.5},\ldots,10^3\}$, by exact leave-one-out cross-validation over training looks. At inference, predictions are converted back to slider values and clipped to $[-1,1]$ before LUT generation. Original training uses 4,500 FiveK Expert C photographs and 8,875 PPR10K Expert A photographs, with 500 and 2,286 test photographs and expert-retouched targets.

\subsection{Method Selection and Availability}
\label{sec:supp_other_research}
LUTBench evaluates methods that edit a whole photograph from a text description and run from public code and weights. Methods without a public implementation are reimplemented only when they require no training, which applies to LCIT. AceTone targets text-conditioned grading, and its released PST preview is evaluated with a reference pair, as described in \cref*{sec:supp_method_profiles}. \Cref{tab:supp_availability} lists the releases used for the evaluated methods and groups the related methods that are not evaluated by the reason for exclusion.

As of November 2026, six related methods have no runnable public model. CLIPtone-GO~\cite{merothiya2026cliptonego} and PixTalk~\cite{conde2025pixtalk} have announced but unreleased implementations, and we found no public code or weights for RetouchIQ~\cite{wu2026retouchiq} and StatLUT~\cite{wang2026statlut}. InstantRetouch~\cite{wu2026instantretouch} provides inference code without a checkpoint, and GenColor~\cite{gencolor2025} releases its texture-preservation network without the color-enhancement model. Each of these methods requires training, so the reimplementation criterion does not cover them. StatLUT also depends on an unreleased collection of 10,000 training LUTs.

JarvisArt~\cite{lin2025jarvisart} and JarvisEvo~\cite{jarvisevo2026} render edits through Adobe Lightroom Classic, which is proprietary software outside the open-source pipeline shared by the evaluated methods. The PerTouch~\cite{pertouch2026} agent requires a user-specified VLM service, which the release does not identify, and an access-restricted SAM3 checkpoint. T2ONet~\cite{shi2021t2onet} encodes requests with an LSTM over a closed vocabulary learned from its MA5k-Req and GIER~\cite{shi2020gier} training requests, so it cannot represent clause terms outside those requests.

SA-LUT~\cite{gong2025salut}, ICTone~\cite{deng2026ictone}, INRetouch~\cite{elezabi2026inretouch}, Deep Preset~\cite{ho2021deeppreset} and Personalized Image Filter~\cite{zhu2025pif} define the look through reference images, which places them outside text-driven look transfer. EyeControl~\cite{eyecontrol2026} targets visual-focus enhancement and conditions on a local instruction and a spatial focus target in addition to a global instruction. ColorCtrl~\cite{yin2026colorctrl} changes the color or material of an object named in the prompt. Both methods address local edits rather than a global photographic look.

\begin{table*}[t]
\centering
\scriptsize
\setlength{\tabcolsep}{3pt}
\renewcommand{\arraystretch}{1.05}
\caption{Implementation and data availability of the evaluated methods and of related methods that LUTBench does not evaluate. Headings of the related-method groups give the reason for exclusion. The data column distinguishes source datasets from exact evaluation material. NI denotes material not identified in the inspected release as of November 2026. A dash indicates no separate dataset entry. Work names link to the cited release or paper.}
\label{tab:supp_availability}
\begin{tabularx}{\textwidth}{@{}p{0.18\textwidth}p{0.27\textwidth}X@{}}
\toprule
Work & Implementation and Model & Data and Evaluation Material \\
\midrule
\multicolumn{3}{@{}l}{\textit{Evaluated Methods}} \\
\href{https://github.com/hmin970922/CLIPtone}{CLIPtone}~\cite{lee2024cliptone} & Code and adapter weights & FiveK and Color Names. Nine-filter rendered pairs: NI. \\
\href{https://library.imaging.org/cic/articles/33/1/38}{LCIT}~\cite{mayer2025lcit} & Paper-based implementation. Author code: NI. & Kodak images and prompt examples. \\
\href{https://github.com/cyclomon/CLIPstyler}{CLIPstyler}~\cite{kwon2022clipstyler} & Code and example decoders & DIV2K, CelebA-HQ and repository examples. \\
\href{https://github.com/sertiscorp/FastCLIPstyler}{FastCLIPstyler}~\cite{suresh2024fastclipstyler} & Code and weights & Painter by Numbers, DTD and ArtEmis. Exact evaluation data: NI. \\
\href{https://github.com/zipengxuc/SpectralCLIP}{SpectralCLIP}~\cite{xu2024spectralclip} & Code using pretrained CLIP & COCO images and style names. Exact pair assignments: NI. \\
\href{https://github.com/YSerin/ZeCon}{ZeCon}~\cite{yang2023zecon} & Code and public diffusion weights & Source datasets and examples. Exact evaluation sample list: NI. \\
\href{https://github.com/gongms00/SPL}{SPL}~\cite{gong2026spl} & Code and LCM weights & PIE-Bench, AnyEdit, FiveK and DPST. Curated subsets: NI. \\
\href{https://github.com/timothybrooks/instruct-pix2pix}{InstructPix2Pix}~\cite{brooks2023instructpix2pix} & Code and weights & Filtered synthetic training data. Exact 2,000 evaluation edits: NI. \\
\href{https://github.com/bytedance/SuperEdit}{SuperEdit}~\cite{li2025superedit} & Code and SD1.5 weights & SuperEdit-40K, MultiReward Real-Edit and MagicBrush. \\
\href{https://ali-vilab.github.io/stylebooth-page/}{StyleBooth}~\cite{han2024stylebooth} & Code and weights & StyleBooth Dataset and 226 Emu Edit style edits. \\
\href{https://huggingface.co/Qwen/Qwen-Image-Edit-2511}{Qwen-Image-Edit-2511}~\cite{wu2025qwenimage} & Weights and inference example & GEdit-Bench and ImgEdit. Checkpoint training data are unspecified. \\
\href{https://github.com/OpenDFM/Image_Edit_Agent}{IEA}~\cite{iea2026} & Code and policy/reward weights & GIER, FiveK, processed annotations and synthetic data. \\
\href{https://github.com/niladridutt/monetGPT}{MonetGPT}~\cite{dutt2025monetgpt} & Code and weights & Generation scripts, FiveK and PPR10K. Exact 400 test IDs: NI. \\
\href{https://github.com/OpenVeraTeam/VeraRetouch}{VeraRetouch}~\cite{veraretouch2026} & Code and weights with renderer & Aether-Bench and AetherRetouch-1M+: NI. \\
\href{https://github.com/martian422/AceTone}{AceTone}~\cite{ma2026acetone} & Code, tokenizer and PST preview. Instruction checkpoint: NI. & Transfer subset. Instruct and AceTone-800K: NI. \\
\href{https://github.com/satoshi-kosugi/PG-IA-NILUT}{PG-IA-NILUT}~\cite{pgianilut2024} & Code and weights & FiveK, PPR10K and test partition files. \\
\midrule
\multicolumn{3}{@{}l}{\textit{Not Evaluated: No Runnable Public Model}} \\
\href{https://github.com/merothiyasatyam/CLIPtone-GO}{CLIPtone-GO}~\cite{merothiya2026cliptonego} & Implementation announced & - \\
PixTalk~\cite{conde2025pixtalk} & Implementation announced & Complete evaluation data: NI. \\
RetouchIQ~\cite{wu2026retouchiq} & Author code and weights: NI. & Complete 300-example RetouchEval set: NI. \\
\href{https://arxiv.org/html/2607.08227v2}{StatLUT}~\cite{wang2026statlut} & Author code and weights: NI. & 10,000-LUT training collection: NI. Tests use PST50 and PhotoNAS. \\
InstantRetouch~\cite{wu2026instantretouch} & Inference code. Checkpoint: NI. & Complete 500-example iRetouch evaluation set: NI. \\
\href{https://github.com/NTUYWANG103/GenColor}{GenColor}~\cite{gencolor2025} & Texture model. Color model announced. & ARTISAN-1.2M 240p preview. \\
\midrule
\multicolumn{3}{@{}l}{\textit{Not Evaluated: Proprietary Software or External Services}} \\
\href{https://github.com/LYL1015/JarvisArt}{JarvisArt} / \href{https://github.com/LYL1015/JarvisEvo}{JarvisEvo}~\cite{lin2025jarvisart,jarvisevo2026} & Code and weights. Rendering requires Adobe Lightroom Classic. & MMArt-Bench, MMArt-PPR10K and ArtEdit-Bench. \\
\href{https://github.com/Auroral703/PerTouch}{PerTouch}~\cite{pertouch2026} & Code and weights. The agent requires a user-specified VLM service and a gated SAM3 checkpoint. & Data-preparation scripts. \\
\midrule
\multicolumn{3}{@{}l}{\textit{Not Evaluated: Closed Request Vocabulary}} \\
\href{https://github.com/jshi31/T2ONet}{T2ONet}~\cite{shi2021t2onet,shi2020gier} & Code and checkpoints. Closed LSTM vocabulary learned from MA5k-Req and GIER requests. & MA5k-Req and GIER. \\
\midrule
\multicolumn{3}{@{}l}{\textit{Not Evaluated: Reference-Guided Input and Not Text-Driven}} \\
\href{https://github.com/Ry3nG/SA-LUT}{SA-LUT}~\cite{gong2025salut} & Code and checkpoint & PST50. \\
\href{https://github.com/dengyuhai/ICTone}{ICTone}~\cite{deng2026ictone} & Author reimplementation and LoRA & TST2K. \\
\href{https://github.com/omarAlezaby/InRetouch}{INRetouch}~\cite{elezabi2026inretouch} & Reference-fitting code & Retouch Transfer Dataset. \\
\href{https://github.com/minhmanho/deep_preset}{Deep Preset}~\cite{ho2021deeppreset} & Code and checkpoints & Unedited training inputs. \\
\href{https://pif.pages.dev/}{Personalized Image Filter}~\cite{zhu2025pif} & Implementation announced & 20 photographic reference sets announced. \\
\midrule
\multicolumn{3}{@{}l}{\textit{Not Evaluated: Local Editing Task}} \\
\href{https://github.com/DragonisCV/EyeControl}{EyeControl}~\cite{eyecontrol2026} & Code. Retouch checkpoint announced. & ControlArt-Bench. \\
\href{https://github.com/zxYin/ColorCtrl_Code}{ColorCtrl}~\cite{yin2026colorctrl} & Code using public diffusion models & Evaluation annotations. \\
\bottomrule
\end{tabularx}
\end{table*}

\section{Additional Results}
\label{sec:supp_results}

\subsection{Sensitivity of the Main Comparisons}
\label{sec:supp_tables}
\Cref{tab:robustness} reports the sensitivity analyses. Across the input-conversion subsets, only MonetGPT, VeraRetouch and AceTone change comparison status, and their CV differences from the unedited input are within 1 $\Delta E_{00}$. The other 19 configurations keep their CV conclusions, including the six supervised configurations with supported CV improvements. Under group resampling, MonetGPT's canonical-error interval includes zero. Under individual-look resampling, the 95\% confidence interval of the canonical-error difference is $[-0.09,0.59]$ for VeraRetouch and $[-1.30,-0.35]$ for AceTone. On Test, 20 of the 21 differences supported in CV remain supported with the same sign, and CLIPtone's difference of $+2.50$ $\Delta E_{00}$ is inconclusive.

\begin{table*}[t]
\centering
\footnotesize
\setlength{\tabcolsep}{2pt}
\caption{Sensitivity of paired differences relative to the unedited input on the corresponding evaluation photograph sets. Positive values indicate higher canonical error. Group CI is the 95\% confidence interval for the difference in canonical error obtained by resampling complete look groups. The CV input-conversion subsets use explicit evidence for LUT input encoding (Evid.), largest mean differences between canonical and alternative target images below 4 $\Delta E_{00}$ (Lower sensitivity), or both conditions (Both criteria). Superscripts $+$/$-$ mark supported differences. Paired differences use unrounded values.}
\label{tab:robustness}
\begin{tabular}{@{}lcccccc@{}}
\toprule
\shortstack{Method minus\\unedited input} & CV & Group CI & Evid. (32) & \shortstack{Lower sensitivity\\(21)} & \shortstack{Both criteria\\(13)} & Test \\
\midrule
CLIPtone~\cite{lee2024cliptone} & $+4.53$$^{+}$ & $[+3.20,+5.91]$ & $+4.51$$^{+}$ & $+4.26$$^{+}$ & $+4.06$$^{+}$ & $+2.50$ \\
LCIT~\cite{mayer2025lcit} & $+1.45$$^{+}$ & $[+1.24,+1.66]$ & $+1.45$$^{+}$ & $+1.55$$^{+}$ & $+1.61$$^{+}$ & $+0.89$$^{+}$ \\
CLIPstyler~\cite{kwon2022clipstyler} & $+13.99$$^{+}$ & $[+12.36,+15.75]$ & $+13.78$$^{+}$ & $+13.72$$^{+}$ & $+13.03$$^{+}$ & $+13.42$$^{+}$ \\
FastCLIPstyler~\cite{suresh2024fastclipstyler} & $+22.99$$^{+}$ & $[+20.20,+26.08]$ & $+22.41$$^{+}$ & $+23.67$$^{+}$ & $+22.65$$^{+}$ & $+21.06$$^{+}$ \\
SpectralCLIP~\cite{xu2024spectralclip} & $+21.55$$^{+}$ & $[+19.89,+23.21]$ & $+21.66$$^{+}$ & $+21.62$$^{+}$ & $+21.93$$^{+}$ & $+20.30$$^{+}$ \\
ZeCon~\cite{yang2023zecon} & $+6.28$$^{+}$ & $[+5.82,+6.67]$ & $+6.30$$^{+}$ & $+6.33$$^{+}$ & $+6.41$$^{+}$ & $+5.46$$^{+}$ \\
SPL~\cite{gong2026spl} & $+10.08$$^{+}$ & $[+8.68,+11.53]$ & $+9.78$$^{+}$ & $+10.09$$^{+}$ & $+9.34$$^{+}$ & $+7.92$$^{+}$ \\
InstructPix2Pix~\cite{brooks2023instructpix2pix} & $+13.55$$^{+}$ & $[+11.95,+14.94]$ & $+13.63$$^{+}$ & $+13.58$$^{+}$ & $+13.80$$^{+}$ & $+12.49$$^{+}$ \\
SuperEdit~\cite{li2025superedit} & $+18.27$$^{+}$ & $[+15.97,+20.53]$ & $+18.40$$^{+}$ & $+19.84$$^{+}$ & $+21.11$$^{+}$ & $+15.82$$^{+}$ \\
StyleBooth~\cite{han2024stylebooth} & $+12.48$$^{+}$ & $[+10.86,+13.92]$ & $+12.62$$^{+}$ & $+12.85$$^{+}$ & $+13.42$$^{+}$ & $+8.91$$^{+}$ \\
Qwen-Image-Edit-2511~\cite{wu2025qwenimage} & $+14.59$$^{+}$ & $[+12.94,+16.48]$ & $+14.30$$^{+}$ & $+14.23$$^{+}$ & $+13.30$$^{+}$ & $+10.81$$^{+}$ \\
IEA~\cite{iea2026} & $+6.13$$^{+}$ & $[+4.92,+7.24]$ & $+6.02$$^{+}$ & $+6.70$$^{+}$ & $+6.76$$^{+}$ & $+6.67$$^{+}$ \\
MonetGPT~\cite{dutt2025monetgpt} & $+0.95$$^{+}$ & $[-0.17,+2.04]$ & $+0.87$ & $+1.47$$^{+}$ & $+1.60$$^{+}$ & $+1.98$$^{+}$ \\
VeraRetouch~\cite{veraretouch2026} & $+0.26$ & $[-0.13,+0.62]$ & $+0.15$ & $+0.50$$^{+}$ & $+0.38$ & $+0.35$ \\
AceTone~\cite{ma2026acetone} & $-0.83$$^{-}$ & $[-1.31,-0.41]$ & $-1.04$$^{-}$ & $-0.51$ & $-0.82$ & $-1.67$$^{-}$ \\
CLIPtone$^*$ & $-2.27$$^{-}$ & $[-3.27,-1.43]$ & $-2.37$$^{-}$ & $-1.90$$^{-}$ & $-1.92$$^{-}$ & $-2.21$$^{-}$ \\
InstructPix2Pix$^*$ & $-2.14$$^{-}$ & $[-2.76,-1.59]$ & $-2.27$$^{-}$ & $-1.63$$^{-}$ & $-1.64$$^{-}$ & $-2.23$$^{-}$ \\
SuperEdit$^*$ & $-2.05$$^{-}$ & $[-2.66,-1.50]$ & $-2.21$$^{-}$ & $-1.45$$^{-}$ & $-1.47$$^{-}$ & $-2.07$$^{-}$ \\
StyleBooth$^*$ & $-2.55$$^{-}$ & $[-3.16,-1.98]$ & $-2.65$$^{-}$ & $-2.12$$^{-}$ & $-2.10$$^{-}$ & $-2.57$$^{-}$ \\
VeraRetouch$^*$ & $-1.41$$^{-}$ & $[-1.90,-0.96]$ & $-1.51$$^{-}$ & $-1.02$$^{-}$ & $-1.03$$^{-}$ & $-1.87$$^{-}$ \\
FastCLIPstyler$^*$ & $+0.96$$^{+}$ & $[+0.24,+1.64]$ & $+0.89$$^{+}$ & $+1.11$$^{+}$ & $+1.02$$^{+}$ & $+1.21$$^{+}$ \\
PG-IA-NILUT$^*$~\cite{pgianilut2024} & $-2.47$$^{-}$ & $[-3.30,-1.69]$ & $-2.55$$^{-}$ & $-2.23$$^{-}$ & $-2.28$$^{-}$ & $-1.93$$^{-}$ \\
\bottomrule
\end{tabular}
\end{table*}

\subsection{Reproduction in the Seen Track}
\label{sec:supp_seen}
All supervised variants have lower canonical error in the Seen track than on Test, as shown in \cref*{tab:main} in the main paper and \cref{tab:supp_seen}. Both settings use the same trained models and photographs. Relative to the unedited input, CLIPtone$^*$, InstructPix2Pix$^*$, SuperEdit$^*$ and StyleBooth$^*$ reduce canonical error by 50--78\% in the Seen track and 20--25\% on Test. These reductions use the corresponding input errors of 7.66 and 10.26 $\Delta E_{00}$ as baselines. For seen looks, paired supervision provides examples of the specific adjustment magnitudes associated with each description. In contrast, transfer to held-out looks requires predicting these magnitudes from descriptions that summarize ranges of adjustments, as detailed in \cref*{sec:supp_descriptions}. The fixed Test partition consists of complete look groups, as described in \cref*{sec:supp_splits}.

\begin{table*}[tp]
\centering\footnotesize
\setlength{\tabcolsep}{3pt}
\renewcommand{\estimateci}[2]{\mbox{#1\,{\scriptsize #2}}}
\caption{Fidelity and paired comparisons in the Seen track on 72 training looks and all 500 test photographs. Brackets give 95\% confidence intervals. Differences are method minus the unedited input or supervised minus the corresponding zero-shot configuration. Superscripts $+$/$-$ mark differences supported by both canonical and minimum error. $^*$ denotes supervised variants. The evaluation scope is defined in \cref*{sec:supp_splits}.}
\label{tab:supp_seen}
\begin{tabular*}{\textwidth}{@{\extracolsep{\fill}}lcccccc@{}}
\toprule
Method & \shortstack{Canonical error\\$\Delta E_{00}\downarrow$} & PSNR $\uparrow$ & SSIM $\uparrow$ & LPIPS $\downarrow$ & \shortstack{Difference from\\unedited input} & \shortstack{Supervised minus\\zero-shot} \\
\midrule
Unedited input & \estimateci{7.66}{[7.13,8.20]} & 23.54 & 0.865 & 0.110 & - & - \\
CLIPtone~\cite{lee2024cliptone} & \estimateci{14.08}{[13.32,14.81]} & 20.81 & 0.777 & 0.195 & \estimateci{$+6.42$$^{+}$}{$[+5.60,+7.24]$} & - \\
LCIT~\cite{mayer2025lcit} & \estimateci{9.08}{[8.62,9.56]} & 23.13 & 0.856 & 0.144 & \estimateci{$+1.42$$^{+}$}{$[+1.31,+1.53]$} & - \\
CLIPstyler~\cite{kwon2022clipstyler} & \estimateci{22.22}{[21.36,23.16]} & 15.43 & 0.428 & 0.447 & \estimateci{$+14.56$$^{+}$}{$[+13.57,+15.67]$} & - \\
FastCLIPstyler~\cite{suresh2024fastclipstyler} & \estimateci{32.18}{[30.65,33.70]} & 11.41 & 0.483 & 0.549 & \estimateci{$+24.53$$^{+}$}{$[+22.66,+26.28]$} & - \\
SpectralCLIP~\cite{xu2024spectralclip} & \estimateci{30.98}{[30.33,31.66]} & 10.68 & 0.225 & 0.580 & \estimateci{$+23.32$$^{+}$}{$[+22.36,+24.25]$} & - \\
ZeCon~\cite{yang2023zecon} & \estimateci{14.32}{[13.84,14.83]} & 17.45 & 0.425 & 0.506 & \estimateci{$+6.67$$^{+}$}{$[+6.26,+7.08]$} & - \\
SPL~\cite{gong2026spl} & \estimateci{18.30}{[17.34,19.32]} & 16.59 & 0.697 & 0.301 & \estimateci{$+10.65$$^{+}$}{$[+9.47,+11.80]$} & - \\
InstructPix2Pix~\cite{brooks2023instructpix2pix} & \estimateci{22.37}{[21.66,23.08]} & 16.51 & 0.597 & 0.354 & \estimateci{$+14.71$$^{+}$}{$[+13.84,+15.55]$} & - \\
SuperEdit~\cite{li2025superedit} & \estimateci{27.72}{[26.66,28.80]} & 12.12 & 0.397 & 0.511 & \estimateci{$+20.06$$^{+}$}{$[+18.85,+21.25]$} & - \\
StyleBooth~\cite{han2024stylebooth} & \estimateci{21.79}{[21.14,22.40]} & 15.04 & 0.543 & 0.420 & \estimateci{$+14.13$$^{+}$}{$[+13.25,+14.99]$} & - \\
Qwen-Image-Edit-2511~\cite{wu2025qwenimage} & \estimateci{20.36}{[19.25,21.50]} & 15.00 & 0.491 & 0.286 & \estimateci{$+12.70$$^{+}$}{$[+11.54,+13.92]$} & - \\
IEA~\cite{iea2026} & \estimateci{17.65}{[17.05,18.31]} & 16.59 & 0.702 & 0.198 & \estimateci{$+10.00$$^{+}$}{$[+9.13,+10.74]$} & - \\
MonetGPT~\cite{dutt2025monetgpt} & \estimateci{11.16}{[10.62,11.75]} & 20.55 & 0.797 & 0.162 & \estimateci{$+3.50$$^{+}$}{$[+2.78,+4.20]$} & - \\
VeraRetouch~\cite{veraretouch2026} & \estimateci{8.29}{[7.87,8.78]} & 23.41 & 0.867 & 0.115 & \estimateci{$+0.64$$^{+}$}{$[+0.30,+1.00]$} & - \\
AceTone~\cite{ma2026acetone} & \estimateci{6.44}{[5.84,7.05]} & 25.42 & 0.910 & 0.079 & \estimateci{$-1.21$$^{-}$}{$[-1.71,-0.65]$} & - \\
CLIPtone$^*$ & \estimateci{1.69}{[1.56,1.81]} & 40.75 & 0.988 & 0.014 & \estimateci{$-5.97$$^{-}$}{$[-6.53,-5.45]$} & \estimateci{$-12.39$$^{-}$}{$[-13.16,-11.60]$} \\
InstructPix2Pix$^*$ & \estimateci{3.77}{[3.67,3.87]} & 29.69 & 0.859 & 0.062 & \estimateci{$-3.89$$^{-}$}{$[-4.41,-3.38]$} & \estimateci{$-18.60$$^{-}$}{$[-19.35,-17.85]$} \\
SuperEdit$^*$ & \estimateci{3.82}{[3.71,3.93]} & 29.86 & 0.857 & 0.065 & \estimateci{$-3.84$$^{-}$}{$[-4.36,-3.33]$} & \estimateci{$-23.89$$^{-}$}{$[-25.03,-22.81]$} \\
StyleBooth$^*$ & \estimateci{3.05}{[2.96,3.14]} & 32.74 & 0.931 & 0.041 & \estimateci{$-4.60$$^{-}$}{$[-5.14,-4.09]$} & \estimateci{$-18.73$$^{-}$}{$[-19.39,-18.07]$} \\
VeraRetouch$^*$ & \estimateci{5.90}{[5.57,6.25]} & 26.04 & 0.908 & 0.066 & \estimateci{$-1.76$$^{-}$}{$[-2.11,-1.42]$} & \estimateci{$-2.39$$^{-}$}{$[-2.77,-2.05]$} \\
FastCLIPstyler$^*$ & \estimateci{8.63}{[8.39,8.87]} & 21.93 & 0.784 & 0.217 & \estimateci{$+0.98$$^{+}$}{$[+0.39,+1.51]$} & \estimateci{$-23.55$$^{-}$}{$[-25.13,-21.98]$} \\
PG-IA-NILUT$^*$~\cite{pgianilut2024} & \estimateci{4.92}{[4.62,5.21]} & 29.48 & 0.905 & 0.069 & \estimateci{$-2.74$$^{-}$}{$[-3.29,-2.21]$} & - \\
\bottomrule
\end{tabular*}
\end{table*}

\subsection{Training Objectives and Strategies}
\label{sec:supp_objectives}
For CLIPtone, the CLIP-objective adapter raises CLIP directional similarity from 0.099 to 0.129 relative to the released model. However, canonical error rises from 13.18 to 16.08 $\Delta E_{00}$, as shown in \cref{tab:supp_objectives}. For full-architecture training, the comparison in \cref{tab:supp_objectives} favors training from scratch over full fine-tuning in CV, with a canonical-error reduction of 1.30. The reduction on Test is 0.52 and is inconclusive under the paired comparison rule.

\begin{table}[t]
\centering\footnotesize
\setlength{\tabcolsep}{2pt}
\caption{CLIPtone~\cite{lee2024cliptone} training configurations on 40 CV looks and 500 validation photographs. Rel. is the released model. Adapter columns update only the text adapter, using target or CLIP supervision. Columns labeled All update both the enhancer and text adapter, from scratch or by fine-tuning. CI rows give the 95\% confidence interval bounds for canonical error. $^*$ marks target supervision. Training settings are in \cref*{sec:supp_method_profiles}.}
\label{tab:supp_objectives}
\begin{tabular*}{\columnwidth}{@{\extracolsep{\fill}}lccccc@{}}
\toprule
 & Rel. & \shortstack{Adapter\\Target$^*$} & \shortstack{Adapter\\CLIP} & \shortstack{All\\Scratch$^*$} & \shortstack{All\\Fine-tune$^*$} \\
\midrule
Canon. $\Delta E_{00}\downarrow$ & 13.18 & 6.81 & 16.08 & 6.38 & 7.69 \\
CI lower & 12.08 & 6.13 & 15.14 & 5.74 & 6.87 \\
CI upper & 14.32 & 7.54 & 16.99 & 7.06 & 8.60 \\
Min. error $\downarrow$ & 12.57 & 6.39 & 15.62 & 6.04 & 6.97 \\
PSNR $\uparrow$ (dB) & 21.64 & 26.17 & 20.10 & 26.85 & 25.19 \\
SSIM $\uparrow$ & 0.773 & 0.907 & 0.728 & 0.925 & 0.902 \\
LPIPS $\downarrow$ & 0.185 & 0.081 & 0.261 & 0.075 & 0.086 \\
\bottomrule
\end{tabular*}
\end{table}

\section{Human Evaluation of Look Descriptions}
\label{sec:supp_human}

\begin{figure*}[t]
\centering
\includegraphics[width=\textwidth]{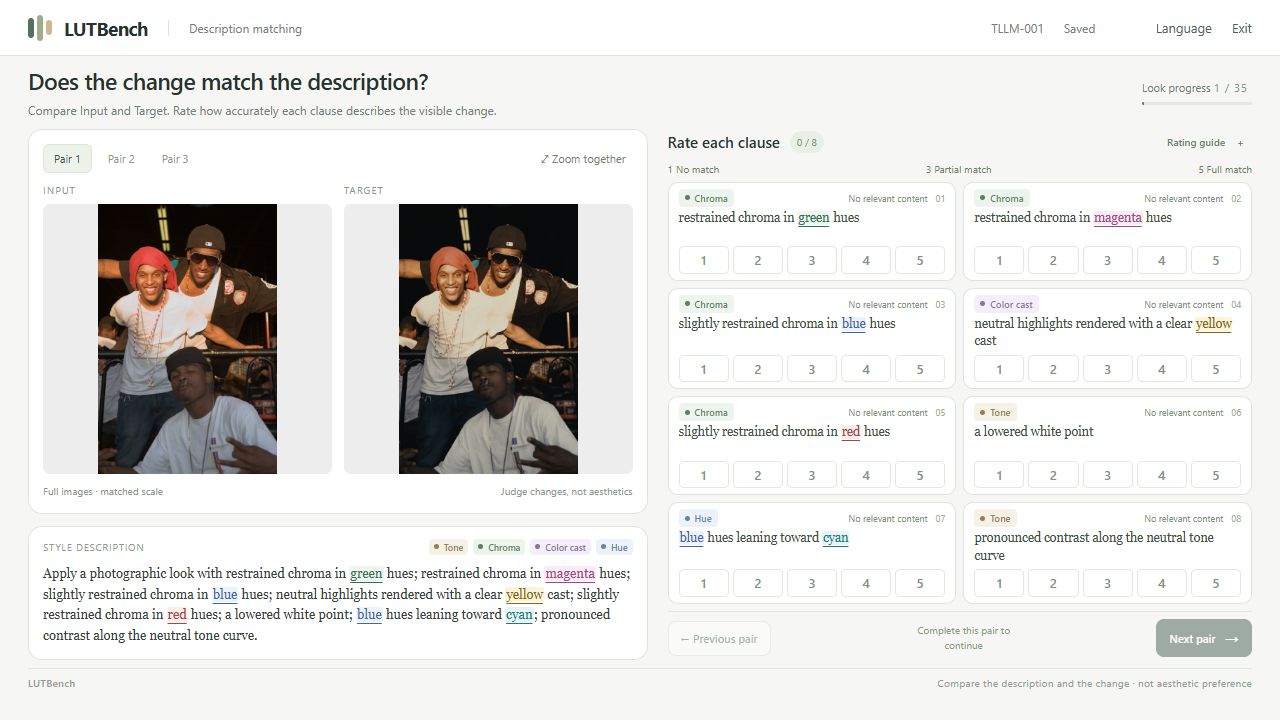}
\caption{Human evaluation interface. Input-target pairs and the full look description appear on the left, with five-point clause ratings and a content-skip option on the right.}
\label{fig:supp_human_interface}
\end{figure*}

\noindent\textbf{Study Objective.}
Human evaluation assessed the visual accuracy of look descriptions against canonical input-target pairs. \Cref{tab:supp_human} summarizes the results.

\noindent\textbf{Materials and Sampling.}
We sampled three photographs for each of the 91 looks from the validation photographs. Photograph identifiers were sorted and permuted with NumPy PCG64 at seed 2027. The first 273 photographs were assigned in groups of three to sorted look identifiers, without reuse across looks. Descriptions, canonical input conversions, image pairs and analysis rules were fixed before rating.

\noindent\textbf{Participants and Assignment.}
We assigned 35 distinct looks and 105 input-target pairs to each of 13 adults who could read English and did not construct the benchmark. The participants were primarily university students, and all had some experience in photography and photo editing. Five independent raters per look completed 455 assignments and 1,365 image-pair assessments. Participants provided informed consent.

\noindent\textbf{Presentation.}
As shown in \cref{fig:supp_human_interface}, each assignment presented a full look description with three full-frame Input/Target pairs, viewed one at a time with Input at left. Look and photograph order were randomized per participant. Clause order followed the look description. Images shared the sRGB display conversion, dimensions and synchronized zoom, without cropping. Participants used a desktop or laptop with night mode and automatic color adjustments disabled. Brief instructions explained tone, chroma, color cast, hue and the response options. Thresholds, support rates, LUT identities and method scores were hidden. Breaks were available every five looks.

\noindent\textbf{Questions and Response Options.}
The screen asked ``Does the change match the description?'' and instructed participants to compare Input and Target and rate how accurately each clause describes the visible change. Each clause had five numbered buttons, with visible anchors ``1 No match'', ``3 Partial match'' and ``5 Full match''. The expandable ``Rating guide'' provided the following definitions:
\begin{itemize}
\item \textbf{1: Does not match at all.} The description does not match the visible change at all.
\item \textbf{2: Mostly does not match.} Most of the description does not match the visible change.
\item \textbf{3: Partly matches.} The description partly matches, with clear discrepancies.
\item \textbf{4: Mostly matches.} The description mostly matches, with only a minor difference in degree.
\item \textbf{5: Fully matches.} Region or color, direction and any stated degree match. Without a degree qualifier, matching region and direction suffices.
\end{itemize}
Scores measure description accuracy. A stronger change receives a higher score only if it better matches the description. A separate ``No relevant content'' option let participants skip clauses whose referenced color or region was absent. This option received no score and required no additional explanation.

\noindent\textbf{Response Inclusion.}
All 13 participants completed their assignments, providing 9,765 responses: 8,208 numeric ratings and 1,557 content skips. The raw counts for scores 1--5 are 45, 619, 1,888, 2,974 and 2,682, respectively.

\noindent\textbf{Evaluation Measures.}
The primary measure is the mean description-match score on the 1--5 scale. Additional measures are the fractions of scores at least four and equal to five, which use numeric responses as denominators, and the content-skip rate, which uses all answered items. For cross-photograph consistency, a unit is one participant's ratings of one clause on the three photographs of a look. A unit is eligible when all three responses are numeric and scores one when all three ratings are at least four. Each measure is computed within each look and then averaged equally across looks with a nonzero denominator. Ordinal Krippendorff's $\alpha$ measures inter-rater agreement on the five-point scale, with each look-photograph-clause combination as an item and content skips treated as missing.

Approximate 95\% confidence intervals for all means and proportions in \cref{tab:supp_human} use the percentile method over 2,000 crossed bootstrap replicates with seed 2027. Each replicate resamples looks and participants independently, weights each rating by the product of their resampling counts and keeps photographs and clauses attached to their looks. Replicates without an eligible denominator are omitted. The interval for $\alpha$ resamples complete look blocks 2,000 times with seed 2027, keeping all items, ratings and missing responses in each block with the rater panel fixed.

\noindent\textbf{Results.}
As shown in \cref{tab:supp_human}, the mean score is close to ``mostly matches'', with 68.94\% of numeric ratings at least four. Chroma (4.19) and hue (4.09) clauses have the highest attribute-level means. Their skip rates of 44.15\% and 33.93\% reflect hues absent from some photographs. Across the eight clause templates, contrast has the highest mean and lowered-white-point clauses the lowest. Across eligible clause-rater units, 45.11\% receive ratings of at least four on all three photographs. Ordinal $\alpha$ is 0.595. Look-level means pool numeric ratings from the five assigned raters, and their crossed bootstrap intervals incorporate variation across looks and raters.

\begin{table}[t]
\centering
\footnotesize
\setlength{\tabcolsep}{3pt}
\caption{Human evaluation of look descriptions. Denom. gives eligible looks/responses, looks/clause-rater units for the three-photograph measure, and items/ratings for $\alpha$. Score measures exclude content skips. Skip rates include all responses.}
\label{tab:supp_human}
\begin{tabularx}{\columnwidth}{@{}Xccc@{}}
\toprule
Measure & Result & 95\% CI & Denom. \\
\midrule
Mean score (1--5) & 3.93 & [3.86,4.00] & 91/8,208 \\
Scores $\geq 4$ (\%) & 68.94 & [65.78,72.10] & 91/8,208 \\
Scores $=5$ (\%) & 32.79 & [29.92,35.93] & 91/8,208 \\
No relevant content (\%) & 15.38 & [13.32,17.36] & 91/9,765 \\
All three scores $\geq 4$ (\%) & 45.11 & [40.33,50.05] & 91/2,262 \\
Ordinal $\alpha$ & 0.595 & [0.568,0.621] & 1,672/8,164 \\
\midrule
\multicolumn{4}{@{}l}{\textit{Mean Score by Attribute (1--5)}} \\
Tone & 3.87 & [3.77,3.99] & 91/3,581 \\
Chroma & 4.19 & [4.09,4.30] & 63/1,301 \\
Color cast & 3.69 & [3.55,3.82] & 78/2,058 \\
Hue & 4.09 & [3.98,4.23] & 81/1,268 \\
\midrule
\multicolumn{4}{@{}l}{\textit{Content-Skip Rate by Attribute (\%)}} \\
Tone & 0.13 & [0.00,0.46] & 91/3,585 \\
Chroma & 44.15 & [38.34,49.78] & 65/2,190 \\
Color cast & 0.67 & [0.00,1.92] & 78/2,070 \\
Hue & 33.93 & [28.67,39.17] & 82/1,920 \\
\midrule
\multicolumn{4}{@{}l}{\textit{Mean Score by Clause Template (1--5)}} \\
Regional tone & 4.04 & [3.91,4.17] & 83/1,932 \\
Lowered white point & 3.11 & [2.94,3.28] & 49/734 \\
Lifted black point & 3.58 & [3.15,3.98] & 25/375 \\
Contrast & 4.47 & [4.34,4.59] & 36/540 \\
Color cast & 3.69 & [3.55,3.82] & 78/2,058 \\
Chroma reduction & 4.18 & [4.07,4.28] & 60/1,260 \\
Chroma enhancement & 4.26 & [3.25,5.00] & 4/41 \\
Hue rotation & 4.09 & [3.98,4.23] & 81/1,268 \\
\bottomrule
\end{tabularx}
\end{table}

\section{Qualitative Comparisons}

\subsection{Look Examples}

\begin{figure*}[p]
\captionsetup{font=small,skip=3pt}
\centering
\includegraphics[width=\textwidth,height=0.90\textheight,keepaspectratio]{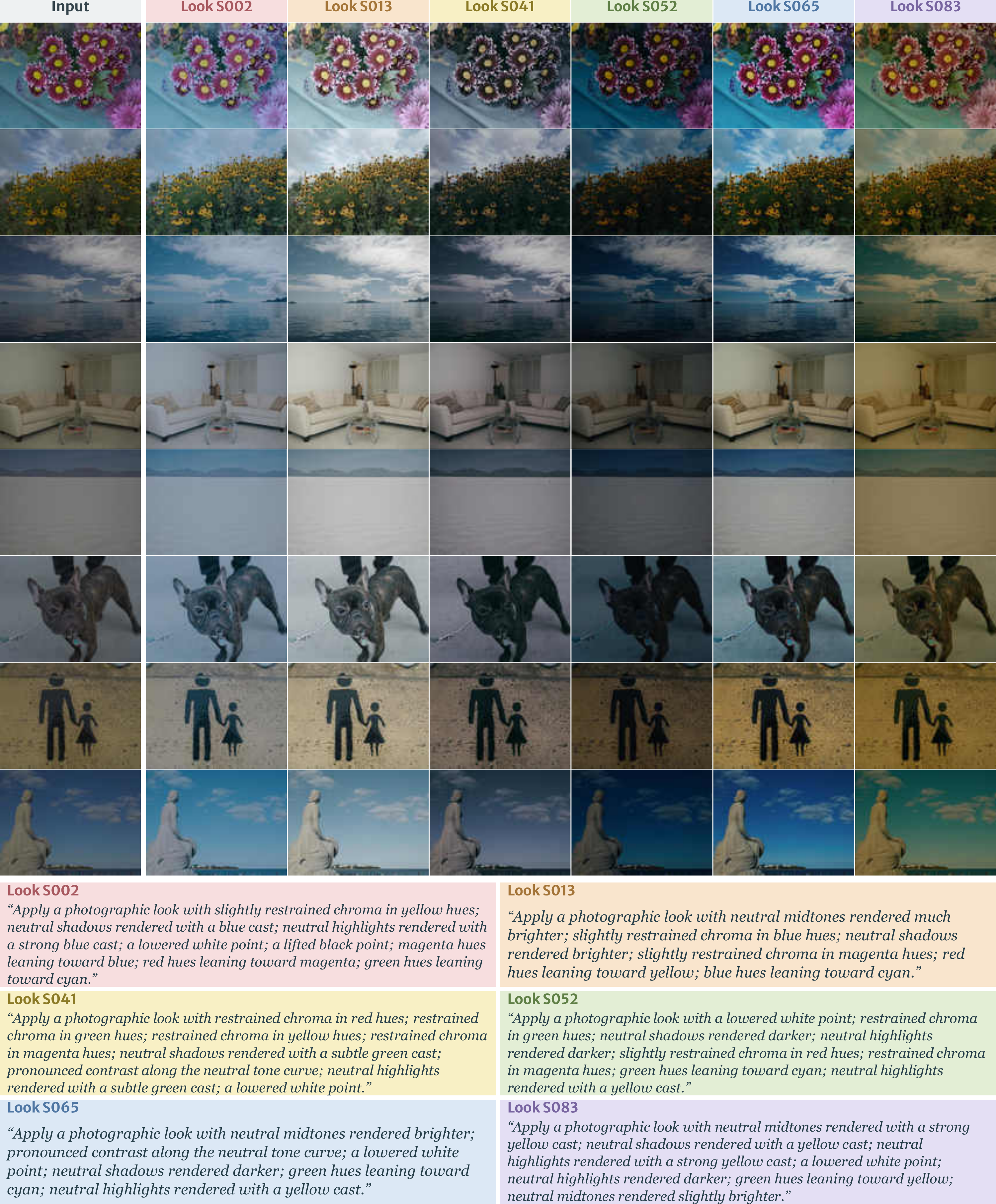}
\caption{Canonical look examples on eight landscape training photographs. Each row shows the input and six LUT-rendered target images, using matched 4:3 crops. Full look descriptions appear below the grid, keyed by color and look identifier.}
\label{fig:supp_looks_landscape}
\end{figure*}

\begin{figure*}[p]
\captionsetup{font=small,skip=3pt}
\centering
\includegraphics[width=\textwidth,height=0.90\textheight,keepaspectratio]{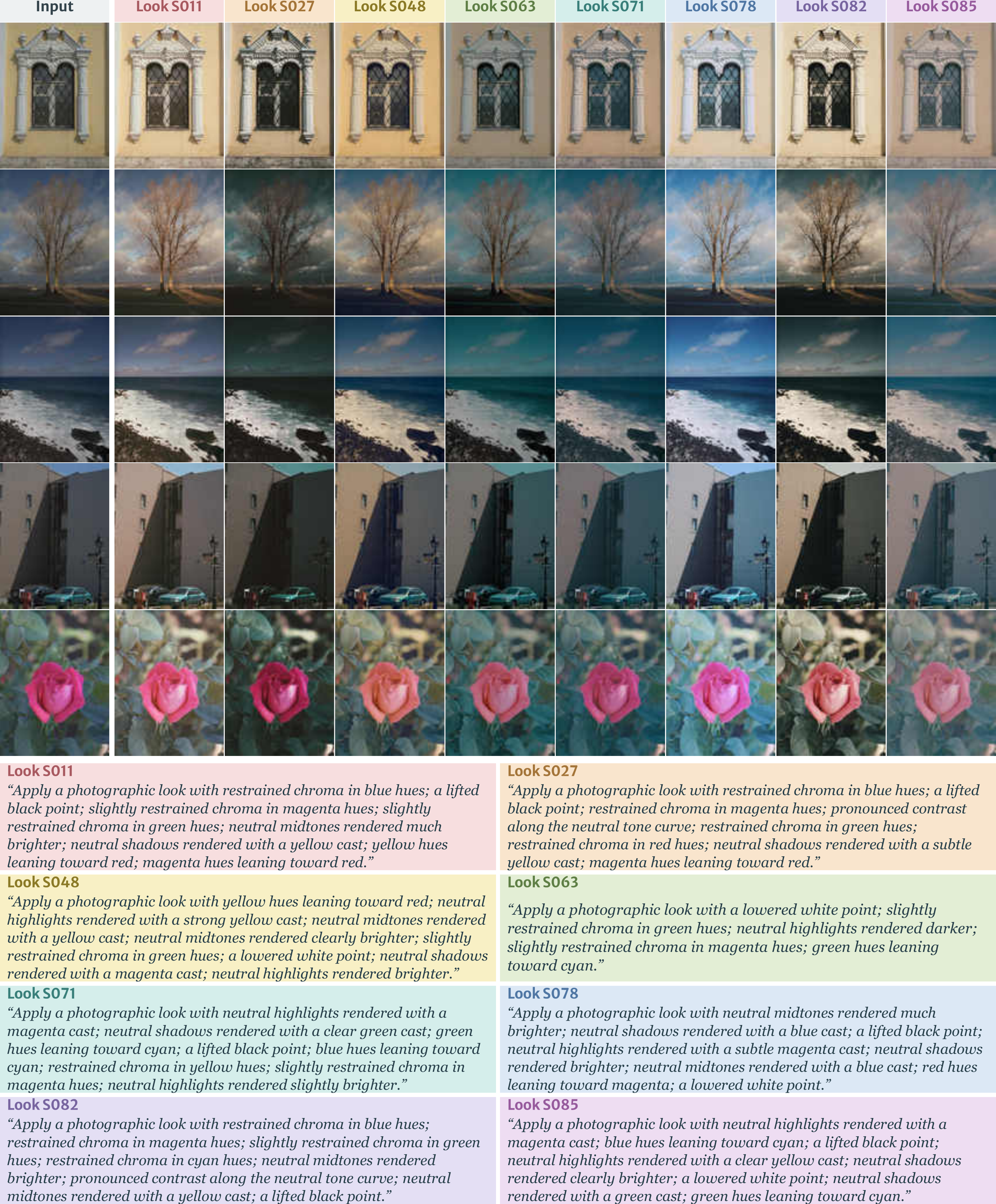}
\caption{Canonical look examples on five portrait-oriented training photographs and eight further looks. Each row uses matched 3:4 crops. Full look descriptions appear below the grid. Both photographs and looks differ from those in \cref{fig:supp_looks_landscape}.}
\label{fig:supp_looks_portrait}
\end{figure*}

\Cref{fig:supp_looks_landscape,fig:supp_looks_portrait} illustrate the look collection through training photographs, their canonical target images and the corresponding look descriptions. Images use the saved sRGB values without additional brightness adjustment.


\subsection{Method Comparisons}
\label{sec:supp_qualitative}
\label{sec:supp_gallery}

\begin{figure*}[tp]
\captionsetup{font=small,skip=3pt}
\centering
\includegraphics[width=\textwidth]{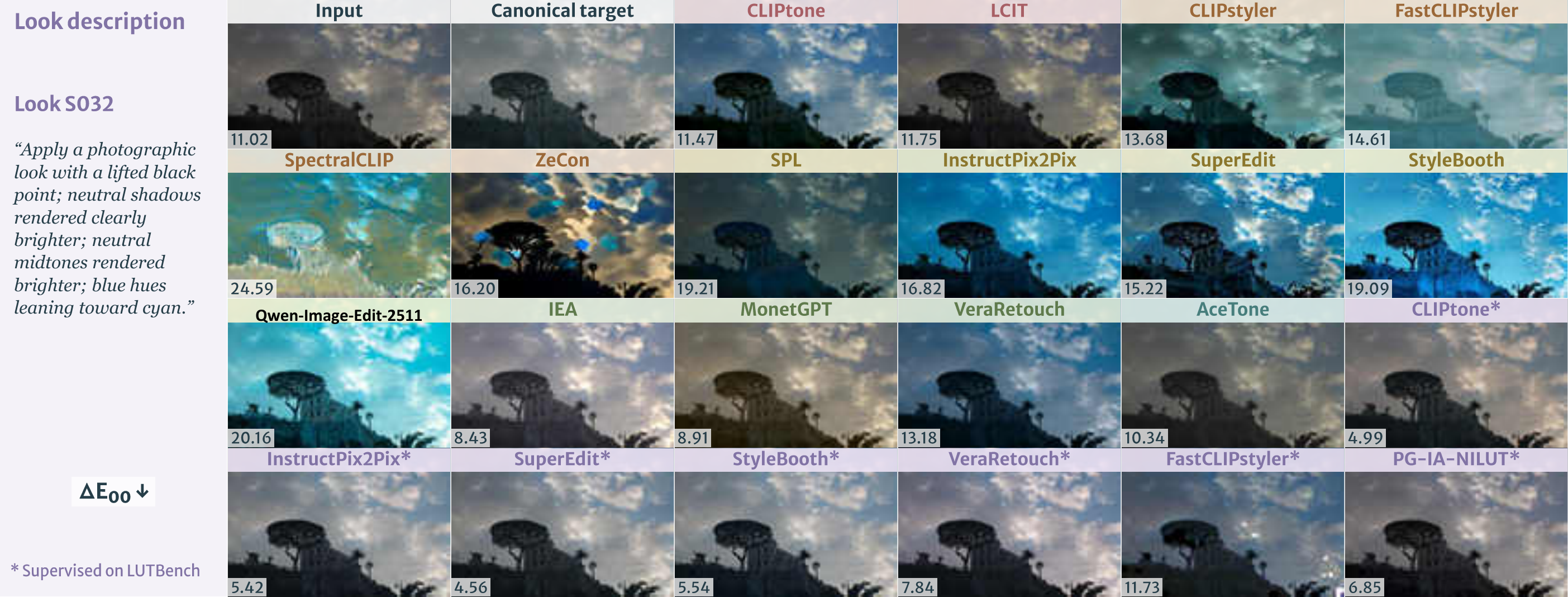}
\caption{Single-photograph comparison under Look S032. The full description appears at left. The grid shows Input, the canonical target and all 22 method configurations.}
\label{fig:supp_single_image}
\end{figure*}

\begin{figure*}[tp]
\captionsetup{font=small,skip=3pt}
\centering
\includegraphics[width=\textwidth]{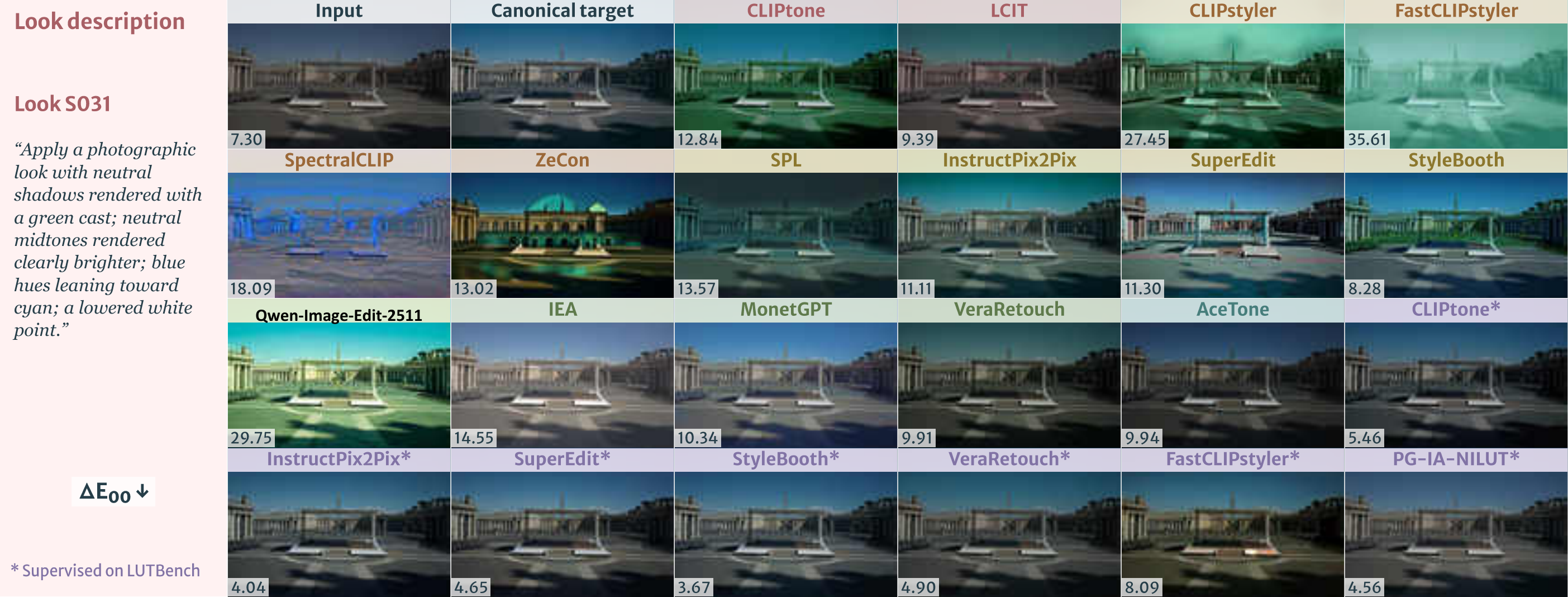}
\caption{Additional qualitative comparison under Look S031. Display conventions follow \cref{fig:supp_single_image}.}
\label{fig:supp_qualitative}
\end{figure*}

\begin{figure*}[tp]
\captionsetup{font=small,skip=3pt}
\centering
\includegraphics[width=\textwidth]{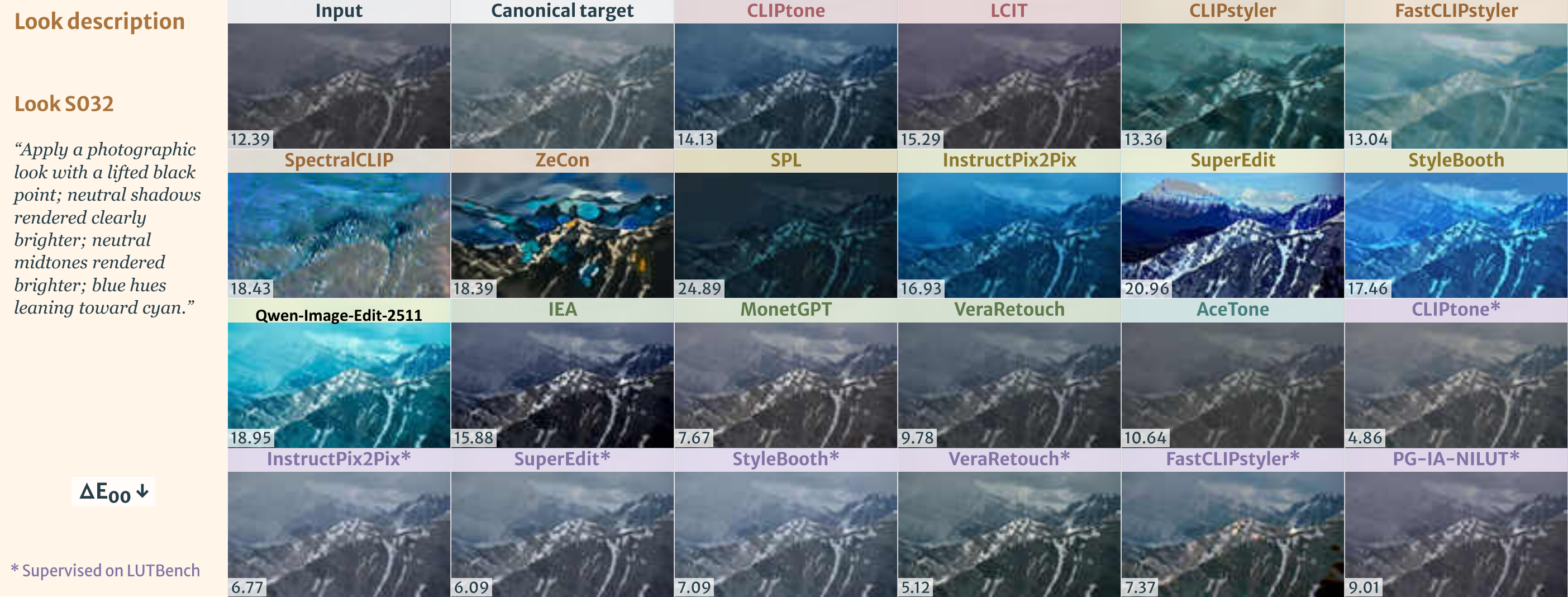}
\caption{Additional qualitative comparison under Look S032. Display conventions follow \cref{fig:supp_single_image}.}
\label{fig:supp_qualitative2}
\end{figure*}

\begin{figure*}[tp]
\captionsetup{font=small,skip=3pt}
\centering
\includegraphics[width=\textwidth]{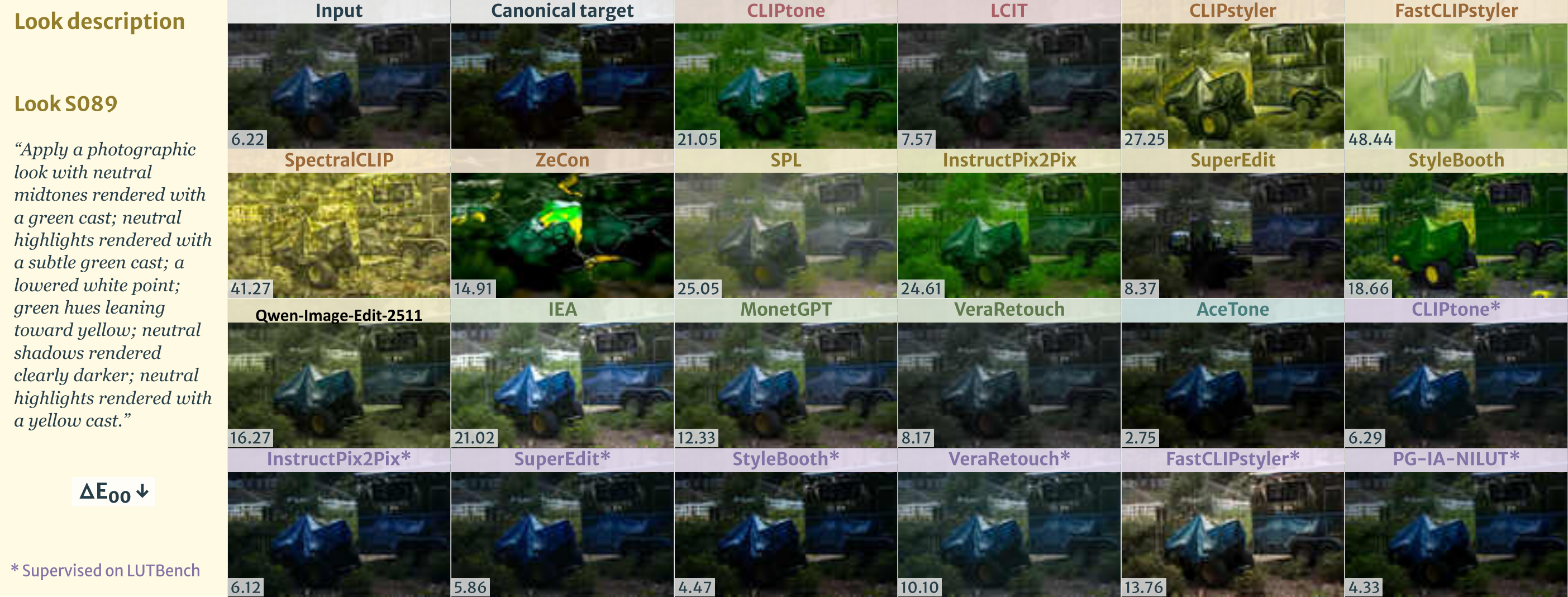}
\caption{Additional qualitative comparison under Look S089. Display conventions follow \cref{fig:supp_single_image}.}
\label{fig:supp_qualitative3}
\end{figure*}

\begin{figure*}[tp]
\captionsetup{font=small,skip=3pt}
\centering
\includegraphics[width=\textwidth]{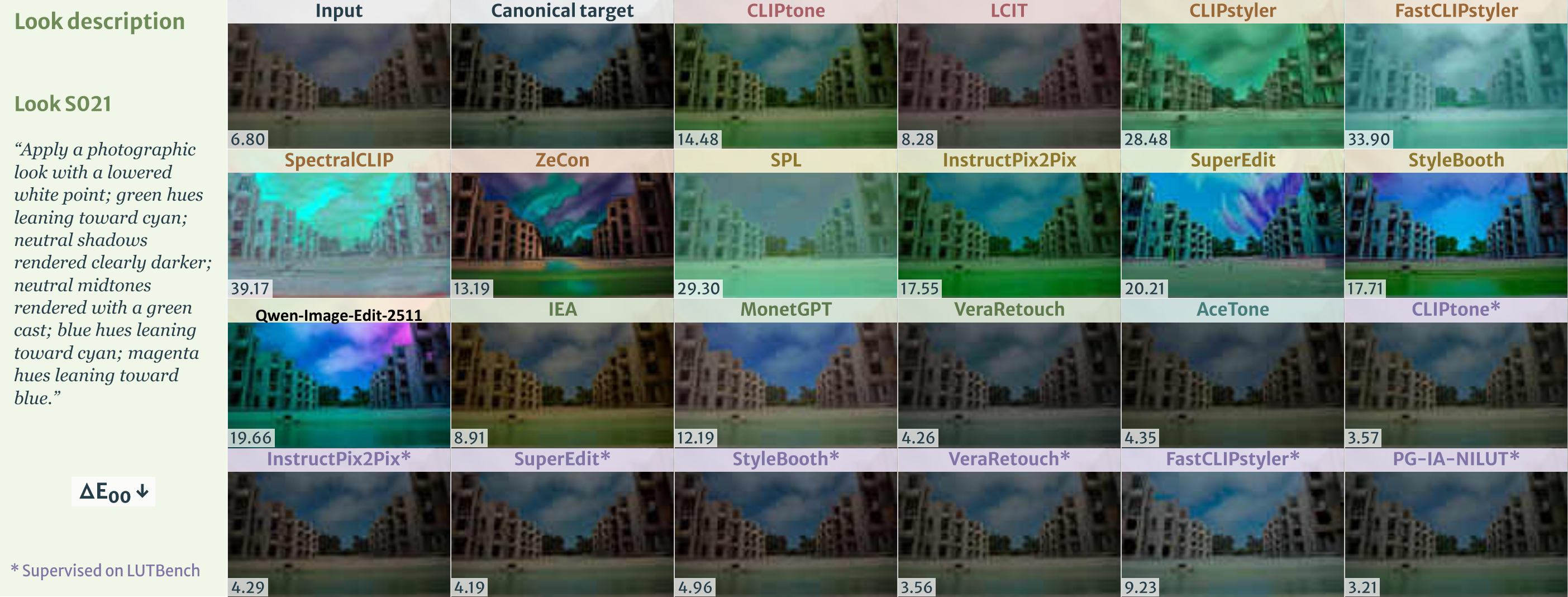}
\caption{Additional qualitative comparison under Look S021. Display conventions follow \cref{fig:supp_single_image}.}
\label{fig:supp_qualitative4}
\end{figure*}

\begin{figure*}[tp]
\captionsetup{font=small,skip=3pt}
\centering
\includegraphics[width=\textwidth]{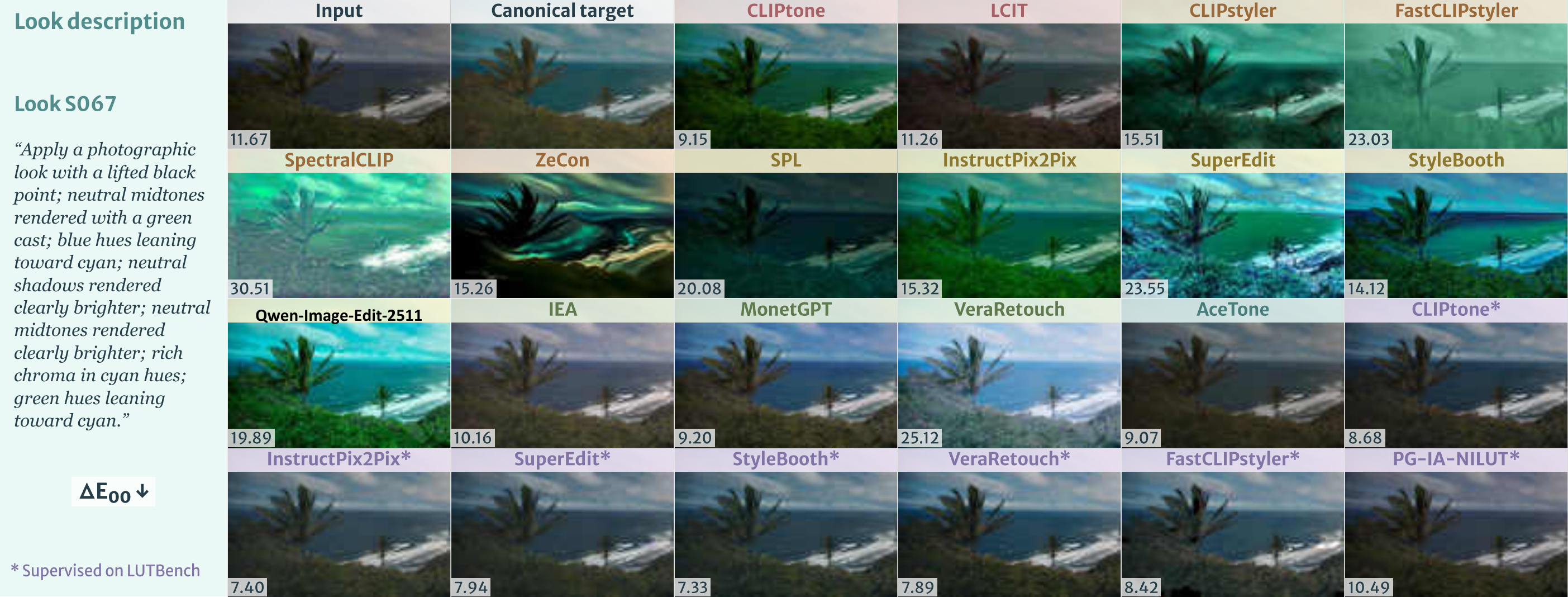}
\caption{Additional qualitative comparison under Look S067. Display conventions follow \cref{fig:supp_single_image}.}
\label{fig:supp_qualitative5}
\end{figure*}

\begin{figure*}[tp]
\captionsetup{font=small,skip=3pt}
\centering
\includegraphics[width=\textwidth]{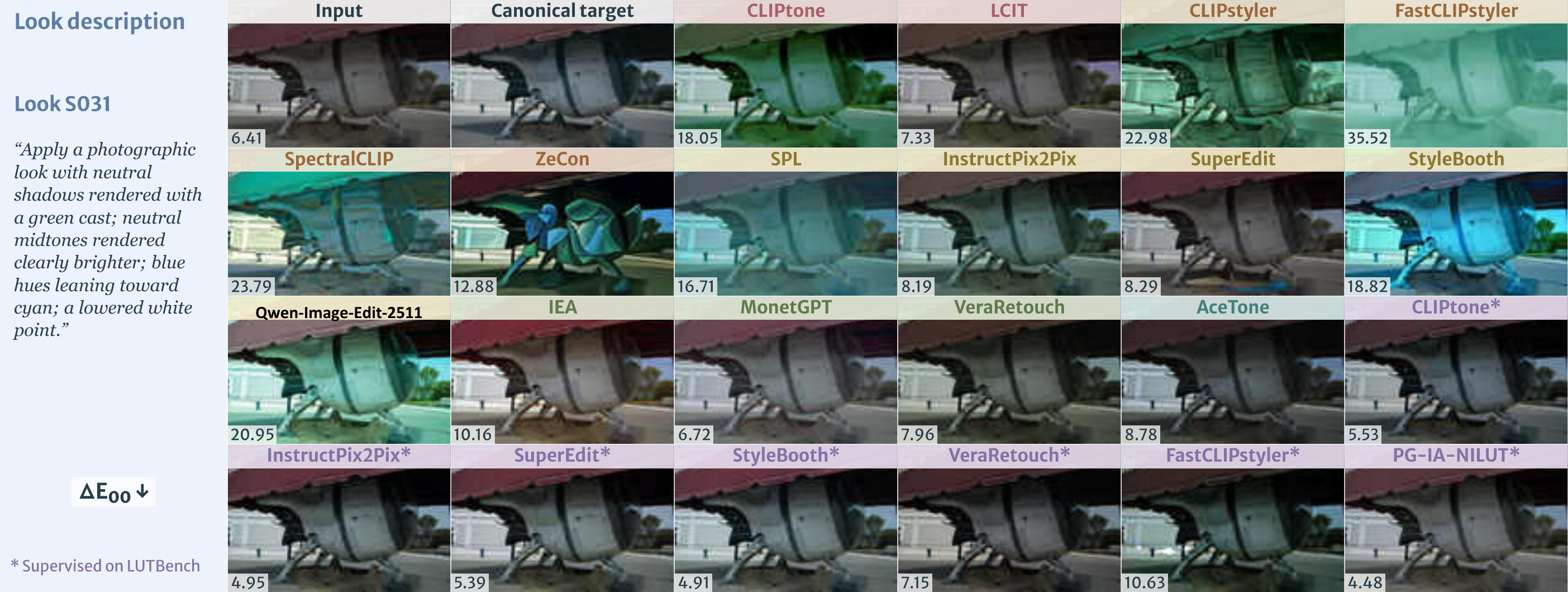}
\caption{Additional qualitative comparison under Look S031. Display conventions follow \cref{fig:supp_single_image}.}
\label{fig:supp_qualitative6}
\end{figure*}

\begin{figure*}[tp]
\captionsetup{font=small,skip=3pt}
\centering
\includegraphics[width=\textwidth]{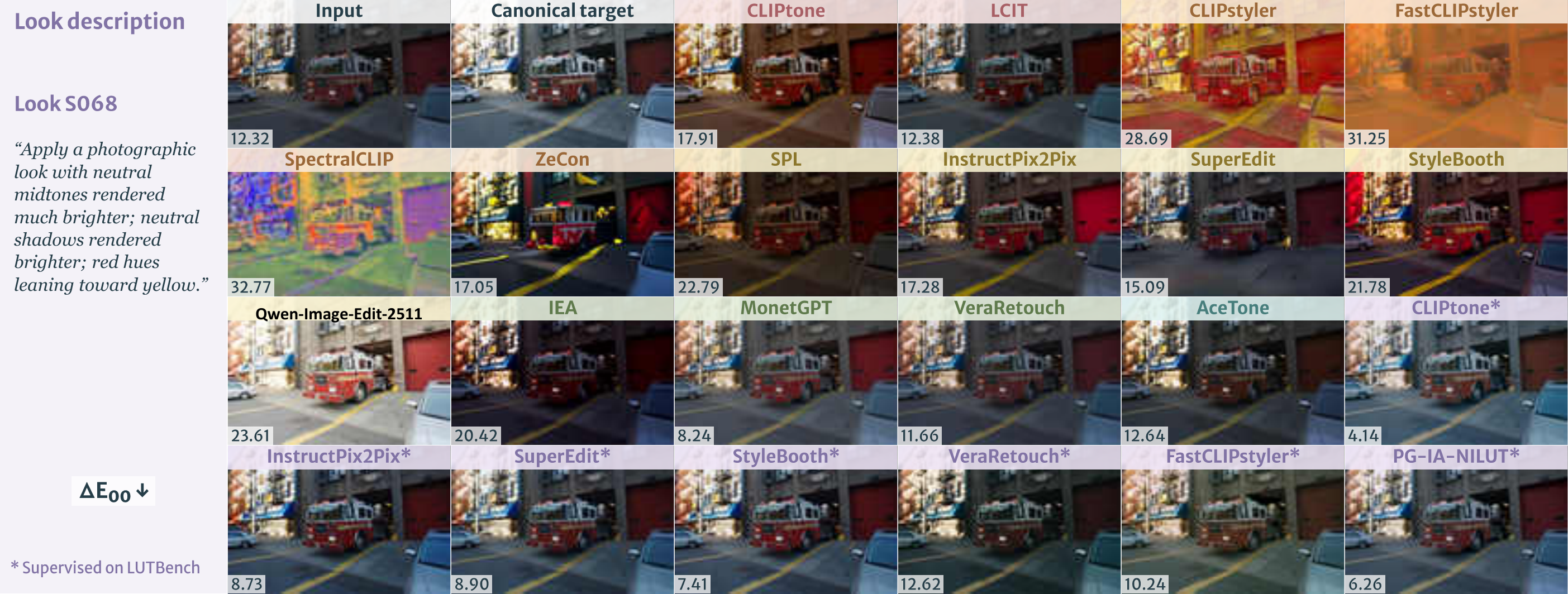}
\caption{Additional qualitative comparison under Look S068. Display conventions follow \cref{fig:supp_single_image}.}
\label{fig:supp_qualitative7}
\end{figure*}

\Crefrange{fig:supp_single_image}{fig:supp_qualitative7} extend \cref*{fig:qualitative} in the main paper with eight randomly selected photograph-look pairs, drawn from validation photographs and held-out CV looks. \Cref{fig:supp_qualitative5,fig:supp_qualitative7} reuse the coast and fire-engine photographs from \cref*{fig:qualitative} under Looks S067 and S068, respectively.

Numerical overlays give the mean pixelwise $\Delta E_{00}\downarrow$ between each output and its canonical target at the saved image resolution. The Input value is the error of the unedited input. Asterisks mark configurations trained on LUTBench, and AceTone uses a reference pair.

Across the eight comparisons, a supervised variant gives the lowest error in seven, and AceTone gives the lowest error under Look S089, with 2.75 against 6.22 for the unedited input. Text style transfer gives the largest error in seven comparisons through painterly textures and broad color washes, reaching 48.44 for FastCLIPstyler under Look S089. Retouching agents stay closer to the input, but their errors vary across photographs. VeraRetouch reaches 4.26 under Look S021, below 6.80 for the unedited input, and 25.12 under Look S067, against 11.67. MonetGPT falls below the unedited input in four comparisons, although its CV mean in \cref*{tab:main} in the main paper is higher than that of the unedited input. The same description can also produce different deviations on different photographs. Under Look S031, zero-shot StyleBooth has an error of 8.28 in \cref{fig:supp_qualitative} and 18.82 in \cref{fig:supp_qualitative6}. These single-photograph examples illustrate method behavior, and the look-level comparisons are in \cref*{tab:main}.


\end{document}